\documentclass{acmtog}
\usepackage{microtype}
\usepackage[linesnumbered, lined, ruled]{algorithm2e}
\usepackage{booktabs}
\usepackage{graphicx}
\usepackage{color}
\usepackage{hyperref}
\usepackage{epsfig}

\acmVolume{VV}
\acmNumber{N}
\acmYear{YYYY}
\acmMonth{Month}
\acmArticleNum{XXX}
\acmdoi{10.1145/XXXXXXX.YYYYYYY}

\begin{document}

\markboth{Z. Yan, H. Zhang, B. Wang, S. Paris, and Y. Yu}{Automatic Photo Adjustment Using Deep Neural Networks}

\title{Automatic Photo Adjustment Using Deep Neural Networks} % title

\author{Zhicheng Yan
\affil{University of Illinois at Urbana Champaign}
Hao Zhang\dag
\affil{Carnegie Mellon University}
Baoyuan Wang
\affil{Microsoft Research}
Sylvain Paris
\affil{Adobe Research}
Yizhou Yu
\affil{The University of Hong Kong and University of Illinois at Urbana Champaign}
}

\category{I.4.3}{Image Processing and Computer Vision}{Enhancement}
\category{I.4.10}{Image Processing and Computer Vision}{Representation}[Statistical]

%\terms{Experimentation, Human Factors}

\keywords{Color Transforms, Feature Descriptors, Neural Networks, Photo Enhancement}

%\acmformat{Pamplona, V. F., Oliveira, M. M., and Baranoski, G. V. G. 2009. Photorealistic models for pupil light
%reflex and iridal pattern deformation.  {ACM Trans. Graph.} 28, 4, Article 106 (August 2009), 11 pages.\newline  DOI $=$
%10.1145/1559755.1559763\newline http://doi.acm.org/10.1145/1559755.1559763}

% \begin{figure}[ht]
% \centering
% \begin{tabular}{c}
%     % Requires \usepackage{graphicx}
%     \includegraphics[width=1.0\linewidth]{img_colormapping/color_mapping.eps}\\
% \end{tabular}
% \caption{Color mapping curves. \textbf{On the left side (from top to bottom)}, input image, semantic label map and ground truth of local Xpro effect. \textbf{On the right side (from top to bottom)}, color mapping curves of L,a,b color channels for both ground truth and our predicted results across different semantic regions.}
% \label{fig:color_mapping}
% \end{figure}

\maketitle

\begin{bottomstuff}
Authors' email addresses: zyan3@illinois.edu, hao@cs.cmu.edu, baoyuanw@microsoft.com, sparis@adobe.com, yizhouy@acm.org.\\
\dag This work was conducted when Hao Zhang was an intern at Microsoft Research.
\end{bottomstuff}

\begin{abstract}
  Photo retouching enables photographers to invoke dramatic visual impressions by artistically enhancing their photos through stylistic color and tone adjustments. However, it is also a time-consuming and challenging task that requires advanced skills beyond the abilities of casual photographers. Using an automated algorithm is an appealing alternative to manual work but such an algorithm faces many hurdles. Many photographic styles rely on subtle adjustments that depend on the image content and even its semantics. Further, these adjustments are often spatially varying. Because of these characteristics, existing automatic algorithms are still limited and cover only a subset of these challenges. Recently, deep machine learning has shown unique abilities to address hard problems that resisted machine algorithms for long. This motivated us to explore the use of deep learning in the context of photo editing. In this paper, we explain how to formulate the automatic photo adjustment problem in a way suitable for this approach. We also introduce an image descriptor that accounts for the local semantics of an image. Our experiments demonstrate that our deep learning formulation applied using these descriptors successfully capture sophisticated photographic styles. In particular and unlike previous techniques, it can model local adjustments that depend on the image semantics. We show on several examples that this yields results that are qualitatively and quantitatively better than previous work.
\end{abstract}

\begin{figure*}[ht]
\centering
\centerline{\includegraphics[width=1.0\linewidth]{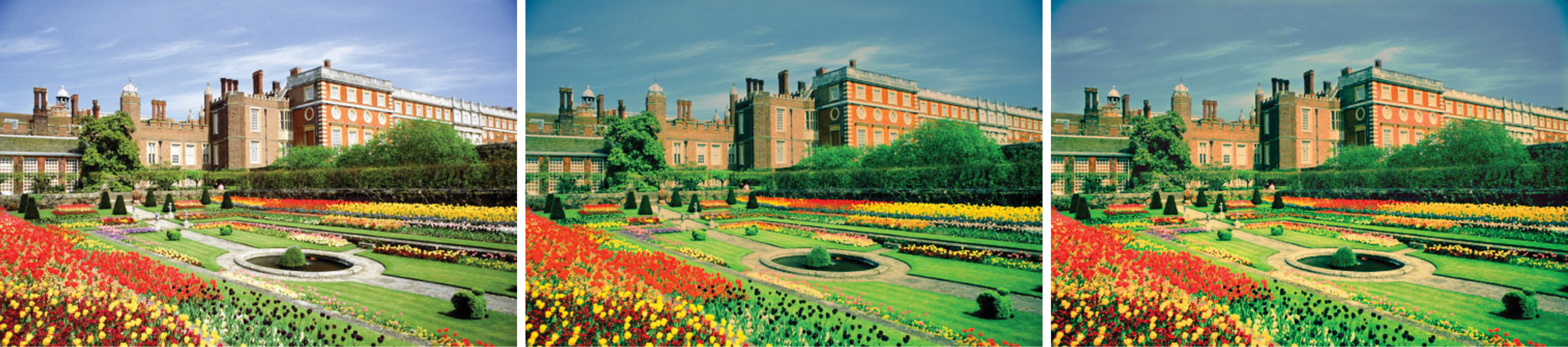}}
  \centerline{\hfill (a) \hfill\hfill (b) \hfill\hfill (c) \hfill}
%\begin{tabular}{c}
%   \includegraphics[width=1.0\linewidth]{images/teaser.pdf}
%   \centerline{\hfill Input Image \hfill\hfill Our Enhanced Result \hfill\hfill Ground Truth \hfill}
%\end{tabular}
   \caption{An example of our semantics-aware photo enhancement style, which extends the ``cross processing" effect in a local manner. \textbf{Left}: input image; \textbf{Middle}: enhanced image by our deep learning based automatic approach; \textbf{Right}: groundtruth image manually enhanced by a photographer, who applied different adjustment parameters in different semantic regions. %The per-pixel L2 regression error in the CIELab space is 8.8.
   See more results of such effects in Section \ref{sec:msr_effect} and the supplemental materials.}
   \label{fig:teaser}
\end{figure*}

\section{Introduction}\label{sec:intro}
With the prevalence of digital imaging devices and social networking, sharing photos through social media has become quite popular. A common practice in this type of photo sharing is artistic enhancement of photos by various Apps such as Instagram. In general, such photo enhancement is artistic because it not only tries to correct photographic defects (under/over exposure, poor contrast, etc.) but also aims to invoke dramatic visual impressions by stylistic or even exaggerated color and tone adjustments. Traditionally, high-quality enhancement is usually hand-crafted by a well-trained artist through extensive labor.

In this work, we study the problem of learning artistic photo enhancement styles from image exemplars. Specifically, given a set of image pairs, each representing a photo before and after pixel-level tone and color enhancement following a particular style, we wish to learn a computational model so that for a novel input photo we can apply the learned model to automatically enhance the photo following the same style.

Learning a high-quality artistic photo enhancement style is challenging for several reasons. First, photo adjustment is often a highly empirical and perceptual process that relates the pixel colors in an enhanced image to the information embedded in the original image in a complicated manner. Learning an enhancement style needs to extract an accurate quantitative relationship underlying this process. This quantitative relationship is likely to be complex and highly nonlinear especially when the enhancement style requires spatially varying local adjustments.
%For this reason, the regression function needs to model the relationship between local adjustment parameters and local information from a surrounding neighborhood of every pixel. As feature vectors describing local information are typically high-dimensional, the relationship we need to model are also likely to be complex and highly nonlinear.
It is nontrivial to learn a computational model capable of representing such a complicated relationship accurately,
and large-scale training data is likely to be necessary.
%in a local region and its surroundings
% needs to model the be learned needs to be local as well. In other words, at each pixel the arguments of the function, as represented by a feature vector, should contain a sufficient amount of
%  local information  Because a significant amount of local features are necessary to provide the discriminative power for capturing the complex and highly non-linear artistic enhancements, the feature vector is high-dimensional and as such will likely require large-scale training data.
Therefore, we seek a learning model scalable with respect to both the feature dimension and data size and  efficiently computable with high-dimensional, large-scale data.

Second, an artistic enhancement is typically semantics-aware. An artist does not see individual pixels; instead he/she sees semantically meaningful objects (humans, cars, animals, etc.) and determines the type of adjustments to improve the appearance of the objects. For example, it is likely that an artist pays more attention to improve the appearance of a human figure than a region of sky in the same photo. We would like to incorporate this semantics-awareness in our learning problem. One challenge is the representation of semantic information in learning so that the learned model can perform image adjustments according to the specific content as human artists do.
%Although there has been much research on example-based photo enhancement techniques, most previous work focuses on correcting photographic defects such as exposure and contrast problems and it is nontrivial to make existing techniques sufficiently scalable to deal with the high-dimensional, large-scale data that we face.

We present an automatic photo enhancement method based on deep machine learning.
This approach has recently accumulated impressive successes in domains
such as computer vision and speech analysis for which the semantics of
the data plays a major role, e.g., \cite{Vincent08,NIPS:2012:CNN}.
This motivated us to explore the use of this class of techniques in our context.
To address the challenges mentioned above, we cast exemplar-based photo adjustment as a regression problem, and use a Deep Neural Network (DNN) with multiple hidden layers to represent the highly nonlinear and spatially varying color mapping between input and enhanced images.
% Deep learning methods \cite{Vincent08,NIPS:2012:CNN} have recently proven to be highly effective in a variety of learning tasks, including object recognition and speech recognition. In addition, a
A deep neural network (DNN) is a universal approximator that can represent arbitrarily complex continuous functions~\cite{Hornik:1989:MFN}. It is also a compact model which is readily scalable with respect to high-dimensional, large-scale data.

Feature design is a key issue that can significantly affect the effectiveness of DNN. To make sure the learned color mapping responds to complex color and semantic information, we design informative yet discriminative feature descriptors that serve as the input to the DNN. For each input image pixel, its feature descriptor consists of three components, which reflect respectively the statistical or semantic information at the pixel, contextual, and global levels. The global feature descriptor is based on global image statistics, whereas the context feature descriptor is based on semantic information extracted from a large neighborhood around the pixel. Understanding image semantics has been made possible with recent advances in scene understanding and object detection. We use existing algorithms to annotate all input image pixels and the semantics information from the annotated images are incorporated into a novel context feature descriptor.

%A practical issue in learning photo enhancement styles is how to select a representative set of images for creating exemplar image pairs. The approach taken in previous work~\cite{Bychkovsky:2011:LPG}, which typically relies on sensor placement to select images based on global statistics, does not work well for our local, semantics-aware enhancement styles. To address this issue we have developed an effective tool for selecting images by minimizing the cross-entropy of an accumulated histogram as defined by a codebook of bundled features, including the local contextual features.

%We have tested our approach on a variety of datasets and found that it produced favorable results when compared with existing techniques. The datasets used in our experiments include artistic effect datasets generated by ¡°Instagram¡± and artistic effect datasets manually created by a professional photographer. We also applied our approach to the ¡°MIT-Adobe FiveK¡± dataset and found that our local approach works well for learning global photo adjustment styles.

\textbf{Contributions}. In summary, our proposed photo enhancement technique has the following contributions.

\begin{itemize}
\item It introduces the first automatic photo adjustment framework based on deep neural networks. A variety of normal and artistic photo enhancement styles can be achieved by training a distinct model for each enhancement style. The quality of our results is superior to that of existing methods.

\item Our framework adopts informative yet discriminative image feature descriptors at the pixel, contextual and global levels. Our context descriptor exploits semantic analysis over multiscale spatial pooling regions. It has achieved improved performance over a single pooling region.
% Nevertheless, our exploration in context feature design for automatic photo enhancement has not been exhaustive, there is potential that variants of our context features could perform even better.

\item Our method also includes an effective algorithm for choosing a representative subset of photos from a large collection so that a photo enhancement model trained over the chosen subset can still produce high-quality results on novel testing images.
\end{itemize}

%\paragraph{Contributions}
%{\flushleft $\bullet$}
%{\flushleft $\bullet$}
%{\flushleft $\bullet$}

While a contribution of our work is the application of deep machine learning in a new context, we use a standard learning procedure and do not claim any contribution in the design of the learning algorithm itself. Similarly, while we propose a possible design for semantic context descriptor, and demonstrate its effectiveness, a comprehensive exploration of the design space for such descriptors is beyond the scope of this paper.

\textcolor{black}{Complete source codes and datasets used by our system are publicly available on Github \footnote{https://github.com/stephenyan1984/dl-image-enhance \newline https://github.com/stephenyan1984/cuda\_convnet\_plus}  .}

\section{Related Work}
\label{sec:relatedwork}

Traditional image enhancement rules are primarily determined empirically. There are many software tools to perform fully automatic color correction and tone adjustment, such as Adobe Photoshop, Google Auto Awesome, and Microsoft Office Picture Manager. In addition to these tools, there exists much research on either interactive ~\cite{LischinskiFUS06,AppProp:08} or automatic ~\cite{Twoscale_Tone:06,Cohen-Or:2006:CH} color and tone adjustment. Automatic methods typically operate on the entire image in a global manner without taking image content into consideration. To address this issue, Kaufman {\em et al.}~\shortcite{Kaufman:2012:CAP} introduces an automatic method that first detects semantic content, including faces, sky as well as shadowed salient regions, and then applies a sequence of empirically determined steps for saturation, contrast as well as exposure adjustment. %Although region detection is learning based, those photo enhancement operations are not. %Conceptually, instead of targeting at specific visual effects, most of such work focuses on improving the commonsense ``quality" in terms of exposure, contrast and color.
% Nevertheless, it is incapable of achieving specifically tailored enhancement styles while our method is designed for achieving such styles. Here the measure of success or failure is whether the output image from a technique is visually and numerically close to a ground truth image generated by the artist who designed the effect.
However, the limit of this approach is that output style is hard-coded
in the algorithm and cannot be easily tuned to achieve a desired
style. In comparison and as we shall see, our data-driven approach can
easily be trained to produce a variety of styles.
%
% Also, as mentioned earlier, today's photo enhancement users go after a
% wide variety of visual effects, some of which may alter an original
% photo in a dramatic or exaggerated manner. Therefore, although these
% tools and algorithms work reasonably well in practice, technically,
% they
Further, these techniques rely on a fixed pipeline that is inherently
limited in its ability to achieve user-preferred artistic enhancement
effects, especially the exaggerated and dramatic ones. In practice, a
fixed-pipeline technique works well for a certain class of adjustments
and only produces approximate results for effects outside this
class. For instance, Bae \emph{et al.}~\shortcite{Twoscale_Tone:06}
do well with tonal global transforms but do not model  local edits, and
Kaufman et al.~\shortcite{Kaufman:2012:CAP} perform well on a
predetermined set of semantic categories but does not handle elements
outside this set.  In comparison, deep learning provides a
universal approximator that is trained on a per-style basis, which is
key to the success of our approach.

%either automatic local and global tone style transfer to meet the reference image Bae {\em et al.}~\shortcite{Twoscale_Tone:06} or automatic color adjustment according to the harmonization rules Cohen-Or\{\em et al.}~\shortcite{Cohen-Or:2006:CH}.
%there is much recent work on either interactive local ~\cite{LischinskiFUS06,AppProp:08} or automatic ~\cite{Twoscale_Tone:06,Cohen-Or:2006:CH} color and tone adjustment from the academia.

Another line of research for photo adjustment is primarily data-driven. Learning based image enhancement ~\cite{PersonalizationSB:2010,Joshi:2010:PPE,CollaborativeEnhancement:2011,Bychkovsky:2011:LPG} and image restoration ~\cite{Restoration:2009} have shown promising results and therefore received much attention. Kang {\em et al.}~\shortcite{PersonalizationSB:2010} found that image quality assessment is actually very much personalized, which results in an automatic method for learning individual preferences in global photo adjustment. Bychkovsky {\em et al.}~\shortcite{Bychkovsky:2011:LPG} introduces a method based on Gaussian processes for learning tone mappings according to global image statistics.
%Although the accompanied dataset, ``MIT-Adobe FiveK", has five different effects generated by global adjustments, the paper only reported results on one of the effects (group C). It is still unclear how well their method handles other effects.
Since these methods were designed for global image adjustment, they do not consider local image contexts and cannot produce spatially varying local enhancements. %it would be a challenge to apply them to large-scale training datasets with millions of high dimensional features necessary for learning spatially varying local enhancement models.
Wang {\em et al.}~\shortcite{Wang:2011:EIC} proposes a method based on piecewise approximation for learning color mapping functions from exemplars.
%In the enhancement stage, it locates the most appropriate mapping functions from a mapping tree using local image features extracted from a small window. Hence,
It does not consider semantic or contextual information either. In addition, it is not fully automatic, and relies on interactive soft segmentation. It is infeasible for this technique to automatically enhance a collection of images.
%For example, Wang {\em et al.}~\shortcite{Wang:2011:EIC} took the Kerneled SVM at each intermediate node of the mapping tree, which requires huge resources for both training and predication.
In comparison, this paper proposes a scalable framework for learning user-defined complex enhancement effects from exemplars. It explicitly performs generic image semantic analysis, and its image enhancement models are trained using feature descriptors constructed from semantic analysis results.

Hwang {\em et al.}~\shortcite{Hwang:2012:CAL} proposes a context-aware
local image enhancement technique.
% , and could be the most relevant
% work.
This technique first searches for the most similar images and then the
most similar pixels within them, and finally apply a combination of
the enhancement parameters at the most similar pixels to the
considered pixel in the new test image. With a sufficiently large
image database, this method works well. But in practice,
nearest-neighbor search requires a fairly large training set that is
challenging to create and slow to search, thereby limiting the
scalability of this approach.
% it is very resource demanding to perform nearest image search and
% nearest pixel search for every pixel in every test image, therefore it
% is not scalable.
%As reflected in past work \cite{TorralbaFF08}, non-parametric image retrieval from small collections often produces spurious matches.
Another difference with our approach is that, to locate the most
similar pixels, this method uses low- and mid-level features
(i.e., color and SIFT) whereas we also consider high-level
semantics. %More importantly, KNN based techniques are sensitive to outliers, and are likely to produce artifacts when certain exemplars in the database exhibit inconsistent enhancement operations.
%In contrast, we would like to learn more abstract and noise-resistant models from exemplars. Hence, unlike previous work, we consider high-level semantic features and learn smooth parametric mapping functions for color transforms using powerful multilayer neural networks which have already been proven to be capable of handling high-dimensional and large-scale datasets.
We shall see in the result section that these differences
have a significant impact on the adjustment quality in several cases.

% %It is well-known that they can perform this task better than metric learning adopted in \cite{Hwang:2012:CAL}.
% This is one of the reasons for the popularity of deep learning these days.
% %In summary, the capability of the learning method is very important; otherwise, even with powerful features, it would not be able to learn the nonlinearity very accurately.
%
%~\cite{Wang:2011:EIC} also proposed a learned based approach to approximate the spatial variant non-linear color mapping functions in a piece-wise manner. Hence, the nonlinearity is %achieved solving multiple local linear or low-order polynomial functions. To locate which local function to use for the enhancement, they further adopt a divide an conquer approach to train %a tree structured classifier, and apply the function stored within the leaf node to perform the color mapping. Although impressive results have been shown in their experiments, they took %very simple local color statistical features with insufficient contextual discriminative power which does not necessarily learn the real content aware mapping functions. While in this work, %we explicitly model the local context features via semantic scene parsing and common object detections, and feed the corresponding context feature descriptors into the learning process.
%

\section{A Deep Learning Model}
Let us now discuss how we cast exemplar-based photo adjustment as a regression problem, and how we set up a DNN to solve this regression problem.
A photo enhancement style is represented by a set of exemplar image pairs $\Lambda= \{I^k,J^k\}_{k=1}^m$,  where $I^k$ and $J^k$ are respectively the images before and after enhancement. Our premise is that there exists an intrinsic color mapping function $\mathcal{F}$ that maps each pixel's color in $I^k$ to its corresponding pixel's color in $J^k$ for every $k$. Our goal is to train an approximate function $\tilde{\cal{F}}$ using $\Lambda$ so that $\tilde{\cal{F}}$ may be applied to new images to enhance the same style there. For a pixel $p_i$ in image $I^k$, the value of $\tilde{\cal{F}}$ is simply the color of image $J^k$ at pixel $p_i$, whereas the input of $\tilde{\cal{F}}$ is more complex because $\tilde{\cal{F}}$ depends on not only the color of $p_i$ in $I^k$ but also additional local and global information extracted from $I^k$, thus we formulate $\tilde{\cal{F}}$ as a parametric function $\tilde{\cal{F}}(\Theta,x_{i})$, where $\Theta$ represents the parameters and $x_i$ represents the feature vector at $p_i$ that encompasses the color of $p_i$ in $I^k$ as well as additional local and global information. With this formulation, training the function $\tilde{\cal{F}}$ using $\Lambda$ becomes computing the parameters $\Theta$ from training data $\Lambda$ through nonlinear regression.

\begin{figure}[!t]
\centerline{\includegraphics[width=3.3in]{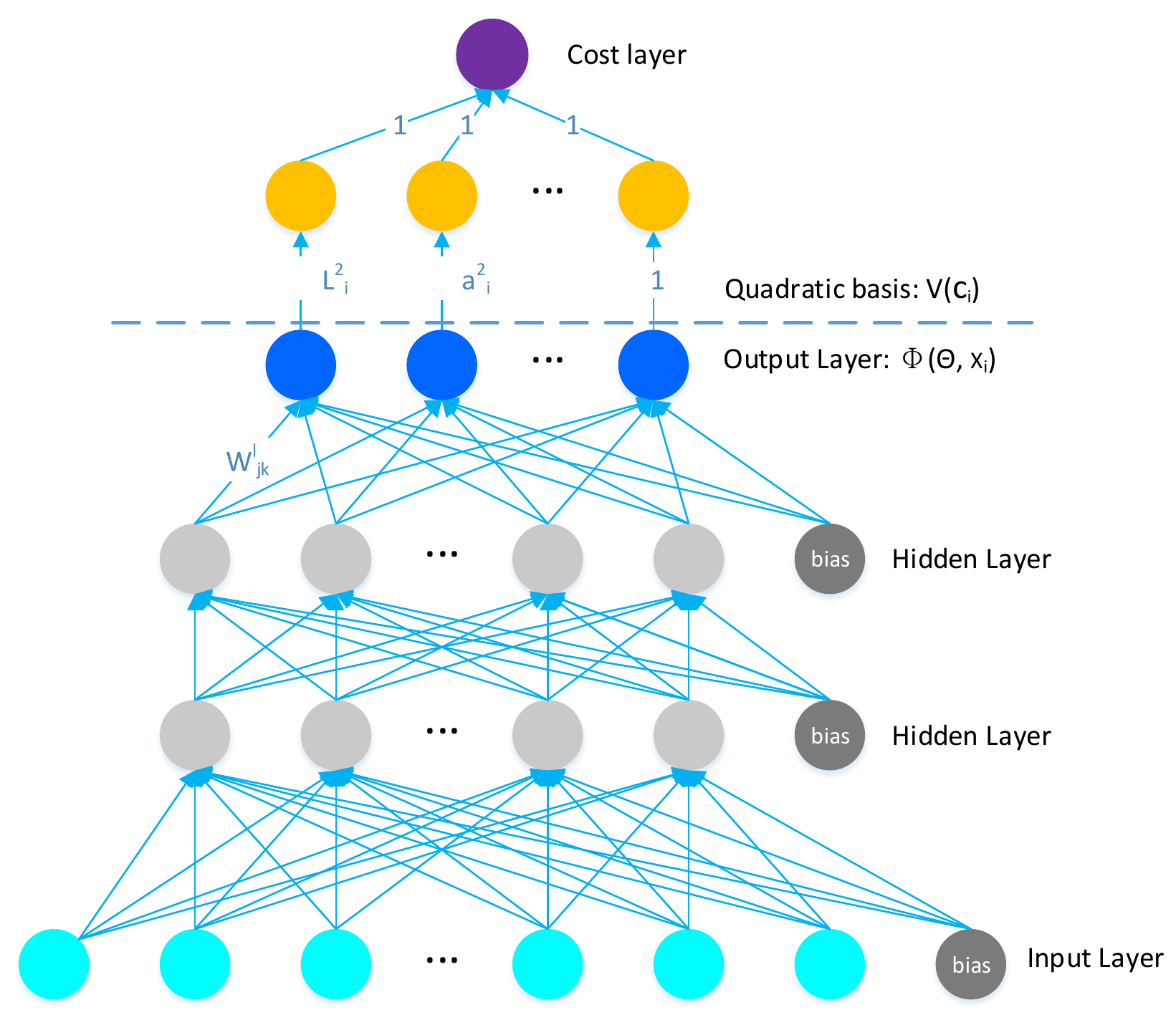}}
  \caption{The architecture of our DNN. The neurons above the dash line indicate how we compute the cost function in (\ref{Equ:lost_function_NN}). Note that the weights for the connections between the blue neurons and the yellow neurons are just the elements of the quadratic color basis, and the activation function in the yellow and purple neurons is the identity function. During training, error backpropagation starts from the output layer, as the connection weights above the dash line have already been fixed.}
  \label{Fig:NN_architecture}
\end{figure}

High-frequency pixelwise color variations are difficult to model because they force us to choose a mapping function which is sensitive to high-frequency details. Such a mapping function often leads to noisy results in relatively smooth regions. To tackle this problem we use a color basis vector $V(c_i)$ at pixel $p_i$ to rewrite $\tilde{\cal{F}}$ as $\tilde{\cal{F}} = \Phi(\Theta, x_{i})V(c_i)$, which expresses the mapped color, $\tilde{\cal{F}}$, as the result of applying the color transform matrix $\Phi(\Theta, x_{i})$ to the color basis vector $V(c_i)$. $V(c_i)$ is a vector function taking different forms when it works with different types of color transforms. In this paper we work in the CIE $Lab$ color space, and the color at $p_i$ is $c_i = [L_i a_i b_i]^T$ and $V(c_i)=[L_i\; a_i\; b_i\; 1]^T$ if we use 3x4 affine color transforms. If we use 3x10 quadratic color transforms, then $V(c_i)=[L^2_{i}\; a_i^2\; b_i^2\; L_ia_i\;  L_ib_i\; a_ib_i\; L_i\; a_i\; b_i\; 1]$. Since the per-pixel color basis vector $V(c_i)$ varies at similar frequencies as pixel colors, it can absorb much high-frequency color variation. By factorizing out the color variation associated with $V(c_i)$, we can let $\Phi(\Theta, x_{i})$ focus on modeling the spatially smooth but otherwise highly nonlinear part of $\tilde{\cal{F}}$.

We learn $\Phi(\Theta,x_{i})$ by solving the following least squares minimization problem defined over all training pixels sampled from $\Lambda$:
\begin{equation}\label{Equ:lost_function}
\arg\min_{\Phi\in\mathcal{H}} \sum_i^n \parallel \Phi(\Theta,x_i) V(c_i) - y_i\parallel^2,
\end{equation}
where $\mathcal{H}$ represents the function space of $\Phi (\Theta, x_i)$ and $n$ is the total number of training pixels. In this paper, we represent $\Phi(\Theta, x_i)$ as a DNN with multiple hidden layers.
%We would like our mapping function $\Phi$ to exhibit the following three properties: 1) capacity to model very complex nonlinearities, 2) smooth function values within local image regions, and 3) fast function evaluation. Inspired by recent advances in deep learning ~\cite{NIPS:2012:CNN}, we propose to adapt feed forward neural networks with multiple hidden layers to our problem, and take such trained neural networks as our mapping functions for color transforms.

\begin{figure*}[!t]
\centerline{\includegraphics[width=1.0\linewidth]{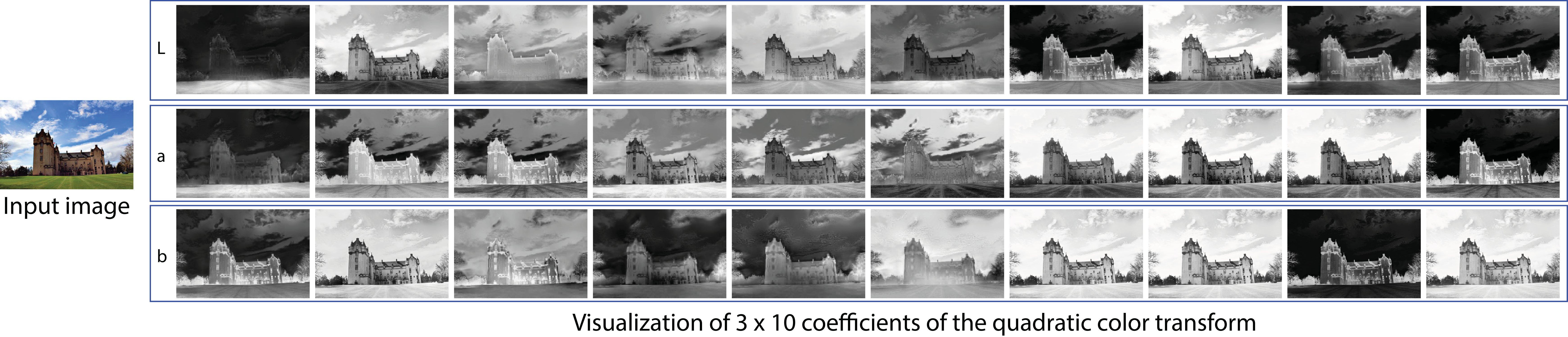}}
\caption{(Left) Input image, and (Right) visualization of its per-pixel quadratic color transforms, $\Phi(\Theta,x_i)$, each of which is a $3\times10$ matrix. Each image on the right visualizes one coefficient in this matrix at all pixel locations. Coefficients are linearly mapped to [0,1] in each visualization image for better contrast.
%The RGB image channels visualize the three elements of the same column in these color transforms.
  This visualization illustrates  two  properties of the quadratic color transforms:
  1)~they are spatially varying and 2)~they are smooth with much high-frequency content suppressed.
%  the high-frequency content of the output coming from that of the input processed by $\Phi$.
%  The first
%   image is the input image. Note that high-frequency details have been
%   suppressed in most of the image regions.
}
\label{fig:vis_color_transform}
\end{figure*}

\subsection{Neural Network Architecture and Training}
Our neural network follows a standard architecture that we describe below for the sake of completeness.

%However, to find $\Phi$, we need to first define the neural network topology, and then determine its weights by minimizing the above loss function (Equation \ref{Equ:lost_function_NN}).
%A multi-layer neural network can be thought of as a parametric model which is capable of approximating an arbitrary multidimensional continuous function.
Multi-layer deep neural networks have proven to be able to represent arbitrarily complex continuous functions~\cite{Hornik:1989:MFN}.
Each network is an acyclic graph, each node of which is a neuron. Neurons are organized in a number of layers, including an input layer, one or more hidden layers, and an output layer. The input layer directly maps to the input feature vector, i.e. $x_{i}$ in our problem. The output layer maps to the elements of the color transform, $\Phi(\Theta,x_{\nu})$. Each neuron within a hidden layer or the output layer takes as input the responses from all the neurons in the preceding layer. Each connection between a pair of neurons is associated with a weight. Let us denote $v_j^l$ as the output of the $j$-th neuron in the $l$-th layer. Then $v_j^l$ is expressed as follows:
\begin{equation}\label{Equ:activation_function}
v_j^l=g\left( w_{j0}^l + \sum_{k>0}w_{jk}^l v_k^{l-1} \right)
\end{equation}
where $w_{jk}^l$ is the weight associated with the connection between the $j$-th neuron in the $l$-layer and the $k$-th neuron in the $(l-1)$-th layer, and $g(z)$ is an activation function which is typically nonlinear. We choose the rectified linear unit (ReLU)~\cite{NIPS:2012:CNN}, $g(z) = \max(0,z)$, as the activation function in our networks. Compared with other widely used activation functions, such as the hyperbolic tangent, $g(z)=\tanh(z)=2/(1+e^{-{2z}})-1$, or the sigmoid, $h(x)=(1+e^{-x})^{-1}$, ReLU has a few advantages, including inducing sparsity in the hidden units and accelerating the convergence of the training process. %What's more important, people found that deep network can be trained very efficiently using ReLU even without pre-training\cite{ReLU:2010}.
Note that there is no nonlinear activation function for neurons in the output layer. The output of a neuron in the output layer is only a linear combination of its inputs from the preceding layer. Figure \ref{Fig:NN_architecture} shows the overall architecture, which has two extra layers (yellow and purple neurons) above the output layer for computing the product between the color transform and the color basis vector.
%and there is no need to perform the nonlinear transformation defined by the activation function. That is to say, $a_j^l=z_j^l=\sum_{k>0}w_{jk}^l*a_k^{l-1}$.
Given a neural network architecture for color mapping, $\mathcal{H}$ in (\ref{Equ:lost_function}) should be the function space spanned by all neural networks with the same architecture but different weight parameters $\Theta$.

%Let us give a concrete example to illustrate how to derive the final output value of a neural network with two hidden layers. For any given input feature vector $x_i$, the output vector $\Phi(w,x_i)$ which is actually our regressed color mapping function, can be expressed as the following closed-form function:
%\begin{equation}
% \Phi(w,x_i)_o = w_{o0}^3 + \sum_{l>0} w_{ol}^3 h(w_{l0}^2 + \sum_{j>0} w_{lj}^2 h(w_{j0}^1 + \sum_{k>0}^D w_{jk}^1x_{ik})) \end{equation}
%where $\Phi(w,x_i)_o$ denotes the $o$-th dimension of vector $\Phi(w,x_i)$, which also maps to the output of the $o$-th neuron in the output layer.

%\subsection{Neural Network Training}
Once the network architecture has been fixed, given a training dataset, we use the classic error backpropagation algorithm to train the weights.
%The main challenge in training a neural network is how to avoid local minima and overfitting. To address those issues,
In addition, we apply the \textit{Dropout} training strategy~\cite{NIPS:2012:CNN,Dropout:2012}, which has been shown very useful for improving the generalization capability. We set the output of each neuron in the hidden layers to zero with probability 0.5. Those neurons that have been ``dropped out" in this way do not contribute to the forward pass and do not participate in error backpropagation. Our experiments show that adding \textit{Dropout} during training typically reduces the relative prediction error on testing data by $2.1\%$, which actually makes a significant difference in the visual quality of the enhanced results.

%In practical applications, a key for success is the design of the feature vector. In our work, the feature vector consists of pixel-wise features, contextual features computed for a local region surrounding the pixel, and global features computed for the underlying image.
Figure \ref{fig:vis_color_transform} visualizes the per-pixel quadratic color transforms, $\Phi(\Theta,x_i)$, generated by a trained DNN for one example image. We can see that the learned color mappings are smooth in most of the local regions.

\section{Feature Descriptors}\label{Sec:feature}
Our feature descriptor $(x_i)$ at a sample pixel $p_i$ serves as the input layer in the neural network. It has three components, $x_i=(x_i^p,\, x_i^c,\, x_i^g)$, where $x_i^p$ represents pixelwise features, $x_i^c$ represents contextual features computed for a local region surrounding $p_i$, and $x_i^g$ represents global features computed for the entire image where $p_i$ belongs.  The details about these three components follow.
\begin{figure}[!t]
\centerline{\includegraphics[width=1.0\linewidth]{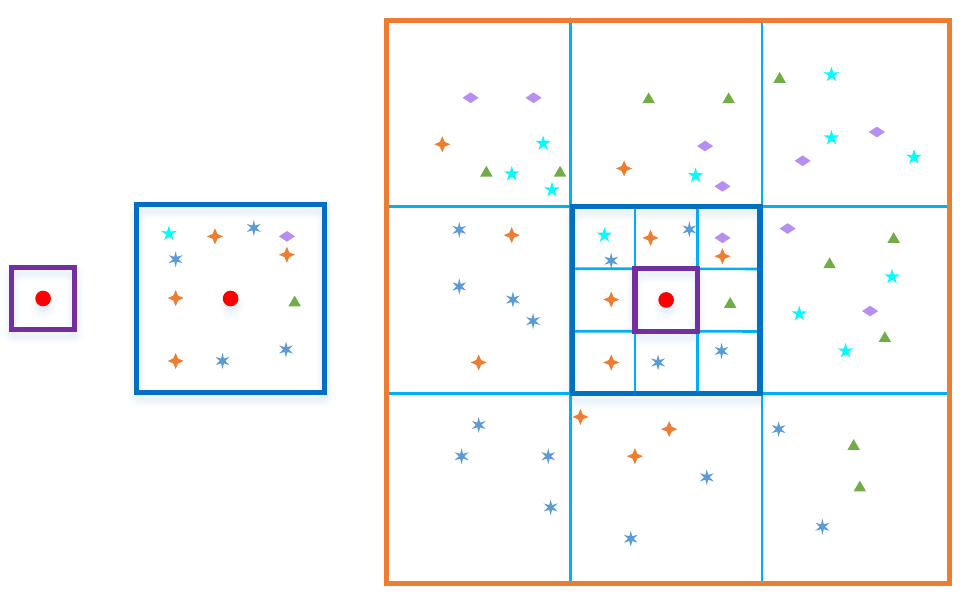}}
  \caption{Our multiscale spatial pooling schema. In each pooling region, we compute a histogram of semantic categories. The shown three-scale scheme has 9*2+1=19 pooling regions. In our experiments, we use a four-scale scheme with 28 pooling regions.}\label{fig:pooling}
\end{figure}

\subsection{Pixelwise Features}
Pixelwise features reflect high-resolution pixel-level image variations, and are indispensable for learning spatially varying photo enhancement models. They are defined as $x_i^p =$ $(c_i, p_i)$, where $c_i$ represents the average color in the CIELab color space within the 3x3 neighborhood, and $p_i=(x_i,y_i)$ denotes the normalized sample position within the image.
%The intuition behind these features is that predicted color transforms in smooth regions should be smooth as well although we prefer spatially varying adjustments.
\begin{figure*}[!t]
  \centerline{\includegraphics[width=1.0\linewidth]{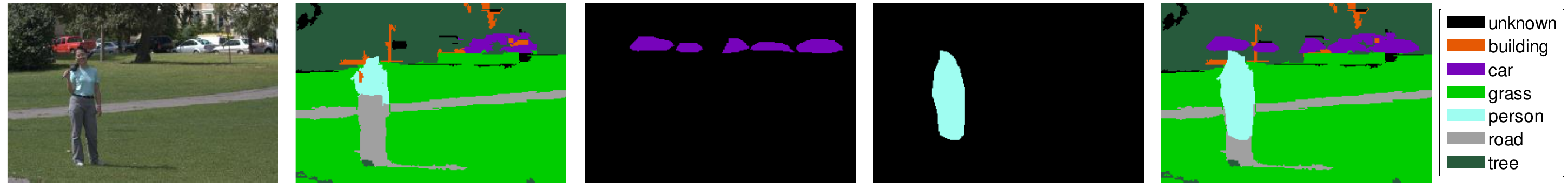}}
  \centerline{\hfill Input Image \hfill\hfill Parsing Map \hfill\hfill Car Detection \hfill\hfill Person Detection \hfill\hfill Final Label Map \hfill\hfill}
  \caption{Pipeline for constructing our semantic label map. Starting with an input image, we first perform scene parsing to obtain a parsing map, then we run object detectors to obtain a detection map for each object category (i.e, car, person), finally we superpose the detection maps onto the parsing map to obtain the final semantic label map (the \textbf{rightmost} image).}\label{fig:map_fusion}
\end{figure*}

\subsection{Global Features}
In photographic practice, global attributes and overall impressions, such as the average intensity of an image, at least have partial influence on artists when they decide how to enhance an image.
%in addition to the local contextual information, the tonal or color adjustment is also frequently influenced by the global scenery distributions and overall impressions.
We therefore incorporate global image features in our feature representation. Specifically, we adopt six types of global features proposed in ~\cite{Bychkovsky:2011:LPG}, including \textit{intensity distribution, scene brightness, equalization curves, detail-weighted equalization curves, highlight clipping, and spatial distribution}, which altogether give rise to a 207-dimensional vector.
%In addition to these hand-crafted features, we optionally use another set of global features, the 4096-dimensional feature vector generated from the second last fully connected layer in a convolutional neural network trained by the ImageNet dataset \cite{NIPS:2012:CNN}, which sometimes can further boost the performance. As the hand-crafted features are low-dimensional and more compact, we use them alone as $x_i^g$ when we need a quick turnaround between training and testing.

\subsection{Contextual Features}
\label{sec:context_ftr}
Our contextual features try to characterize the distribution of semantic categories, such as sky, building, car, person, and tree, in an image. Such features are extracted from semantic analysis results within a local region surrounding the sample pixel. Typical image semantic analysis algorithms include scene parsing~\cite{Tighe:2010:SSN,LiuYT11} and object detection~\cite{ViolaJones01,FMR08,Regionlet:2013:ICCV}. Scene parsing tries to label every pixel in an image with its semantic category. Object detection on the other hand trains one highly specialized detector for every category of objects (such as dogs). Scene parsing is good at labeling categories (such as grass, roads, and sky) that have no characteristic shape but relatively consistent texture.
These categories have a large scale, and typically form the background of an image.
Object detectors are better at locating categories (such as persons and cars), which are better characterized by their overall shape than local appearance.
These categories have a smaller scale, and typically occupy the foreground of an image.
Because these two types of techniques are complementary to each other, we perform semantic analysis using a combination of scene parsing and object detection algorithms. Figure \ref{fig:map_fusion} illustrates one fusion example of the scene parsing and detection results.

We use existing algorithms to automatically annotate all input image pixels and the semantics information from the annotated images are gathered into a novel context feature descriptor.
During pixel annotation, we perform scene parsing using the state-of-the-art algorithm in ~\cite{Tighe:2010:SSN}. \textcolor{black}{The set of semantic categories, $\mathcal{S}^p$, during scene parsing include such object types as sky, road, river, field and grass.
%$\mathcal{S}^p$=$\{$\textit{unknown, building, bus, car, crosswalk, desert, field, grass, mountain, person, plant, river, road, rock, sand, sea, sidewalk, sky, train, tree}$\}$
}. After the scene parsing step, we obtain a parsing map, denoted as $\mathcal{I}^p$, each pixel of which receives one category label from $\mathcal{S}^p$, indicating that with a high probability, the corresponding pixel in the input image is covered by a semantic instance in that category.
%Thanks to the MRF constraint during the parsing process, homogenous neighboring pixels trend to have the same semantic labeling. Therefore, scene parsing not only achieves a rough segmentation, but also obtains pixel wise category annotations. As our scene parsing component mainly deals with outdoor images, to handle indoor scene,
We further apply the state-of-the-art object detector in ~\cite{Regionlet:2013:ICCV} to detect the pixels covered by a predefined set of foreground object types, $\mathcal{O}^d$,
\textcolor{black}{which include person, train, bus and building}.
After the detection step, we obtain one confidence map for each predefined type. We fuse all confidence maps into one by choosing, at every pixel, the object label that has the highest confidence value. This fused detection map is denoted as $\mathcal{I}^d$. We further merge $\mathcal{I}^d$ with $\mathcal{I}^p$ so that those pixel labels from $\mathcal{I}^d$ with confidence larger than a predefined threshold are used to overwrite the corresponding labels from $\mathcal{I}^p$.
\textcolor{black}{
Since scene parsing and object detection results tend to be noisy, we rely on voting and automatic image segmentation to perform label cleanup in the merged label map. Within each image segment, we reset the label at every pixel to the one that appears most frequently in the segment. In our experiments, we adopt the image segmentation algorithm in \cite{arbelaez2011contour}. This cleaned map becomes our final semantic label map, $\mathcal{I}^{label}$}.
%This is to say, our system ends up with $\mathcal{N} = \mathcal{N}^p\bigcup \mathcal{N}^d$ number of category by merging categories from $\mathcal{N}^p$ and $\mathcal{N}^d$ as there are some overlapping between them.
%Latest computer vision algorithms are able to tell with a reasonable accuracy the type of scene where the photo was taken, which types of objects and surfaces are present in the photo, and the type of the object or surface covering every pixel.

%\begin{figure}[!t]
%  \centering
%  % Requires \usepackage{graphicx}
%  \includegraphics[width=1.0\linewidth]{images/W_without_context}\\
%  \caption{An example of local enhancement. \textbf{Top Left}: Input image; \textbf{Top Right}: enhanced image without using contextual features; \textbf{Bottom Left:} enhanced image using contextual features; \textbf{Bottom Right:} enhanced image by photographer. Note that, without discriminative contextual features, the sky and building regions have similar adjustment parameters. }
%  \label{fig:Compare_context}
%\end{figure}
%

Given the final semantic label map for the entire input image, we construct a contextual feature descriptor for each sample pixel to represent multiscale object distributions in its surroundings. For a sample point $p_i$, we first define a series of nested square regions, $\{R_0, R_1, \ldots, R_{\tau}\}$, all centered at $p_i$. The edge length of these regions follows a geometric series, i.e. $\lambda_k=3\lambda_{k-1}(k=1,\ldots,\tau)$, making our feature representation more sensitive to the semantic contents at nearby locations than those farther away. We further subdivide the ring between every two consecutive squares, $R_{k+1}-R_k$, into eight rectangles, as shown in Figure \ref{fig:pooling}. Thus, we end up with a total of $9\tau+1$ regions, including both the original regions in the series as well as regions generated by subdivision. For each of these regions, we compute a semantic label histogram, where the number of bins is equal to the total number of semantic categories, $N=|\mathcal{S}^p \bigcup \mathcal{O}^d|$. Note that the histogram for $R_k$ is the sum of the histograms for the nine smaller regions within $R_k$. Such spatial pooling can make our feature representation more robust and better tolerate local geometric deformations. The final contextual feature descriptor at $p_i$ is defined to be the concatenation of all these semantic label histograms. Our multiscale context descriptor is partially inspired by shape contexts~\cite{Belongie:2002:SMO}. However, unlike the shape context descriptor, our regions and subregions are either rectangles or squares, which facilitate fast histogram computation based on integral images (originally called summed area tables)~\cite{ViolaJones01}. In practice, we pre-compute $N$ integral images, one for each semantic category. Then the value of each histogram bin can be calculated within constant time, which is extremely fast compared with the computation of shape contexts.
To the best of our knowledge, our method is the first one that explicitly constructs semantically meaningful contextual descriptors for learning complex image enhancement models.

It is important to verify whether the complexity of our contextual features is necessary in learning complex spatially varying local adjustment effects. We have compared our results against those obtained without contextual features as well as those obtained from simpler contextual features based on just one pooling region (vs. our 28 multiscale regions) at the same size as our largest region. From Figure \ref{fig:Compare_context}, we can see that our contextual features are able to produce local adjustment results closest to the ground truth.

{\flushleft \textbf{Discussion}}. The addition of this semantic component into our feature vectors is a
major difference with previous work. As shown in
Figure~\ref{fig:Compare_context} and in the result section, the
design that we propose for this component is effective and produces
a significant improvement in practice. That said, we acknowledge that
other options may be possible and we believe that exploring the design
space of  semantic descriptors is an exciting avenue for
future work.

\section{Training Data Sampling and Selection}
%Therefore, we can parameterize $\Phi(x_i)$ as $\Phi(w,x_i)$ , where $x_i$ serves as the input layered (defined in Section \ref{Sec:feature}) while $w$ represents the weight vectors for the neural network. The CMF can therefore be learned by solving the following optimization:
%\begin{equation}\label{Equ:lost_function_NN}
%\Phi = \arg\min_{\Phi\in\mathcal{H}} \sum_i^n \parallel \Phi(w,x_i)^T Q_i - y_i\parallel^2
%\end{equation}

\subsection{Superpixel Based Sampling}
When training a mapping function using a set of images, we prefer not to make use of all the pixels as such a dense sampling would result in unbalanced training data. For example, we could have too many pixels from large \textit{``sky"} regions while relatively few from smaller \textit{``person"} regions, which could eventually result in a serious bias in the trained mapping function. In addition, an overly dense sampling unnecessarily increases the training cost, as we need to handle millions of pixel samples. Therefore, we apply a superpixel based method to collect training samples.
For each training image $I$, we first apply the graph-based segmentation~\cite{Superpixel:2004:EGI} to divide the image into small homogeneous yet irregularly shaped patches, each of which is called a superpixel. Note that a superpixel in a smooth region may be larger than one in a region with more high-frequency details. We require that the color transform returned by our mapping function at the centroid of a superpixel be used for predicting with sufficient accuracy the adjusted color of all pixels within the same superpixel. To avoid bias, we randomly sample a fixed number of pixels from every superpixel. Let $\nu$ be any superpixel from the original images (before adjustment) in $\Lambda$, and $S_{\nu}$ be the set of pixels sampled from $\nu$. We revise the cost function in (\ref{Equ:lost_function}) as follows to reflect our superpixel-based sampling and local smoothness requirement.
%However, $x_i^{\nu}$ does not necessarily and always serve as the representive pixel, we therefore slightly revise the neural network cost function in Equation \ref{Equ:lost_function_NN} to respect a random subset of pixels within $\nu$.
\begin{equation}\label{Equ:lost_function_NN}
 %\Phi = \arg\min_{\Phi\in\mathcal{H}} \sum_i^n \{\parallel \Phi(w,x_i)^T Q_i - y_i\parallel^2 + \sum_{j\in S(x_i^{\nu})}\parallel \Phi(w,x_{j})^T Q_{j} - y_{j}\parallel^2\}
 \sum_{\nu} \sum_{j\in S_{\nu}}\parallel \Phi(\Theta,x_{\nu}) V(c_j) - y_j \parallel^2,
\end{equation}
where $\Theta$ represents the set of trained weights in the neural network, $x_{\nu}$ is the feature vector constructed at the pixel closest to the centroid of $\nu$, $V(c_j)$ denotes the color basis vector of a sample pixel within $\nu$, and $y_j$ denotes the adjusted color of the same sample within $\nu$.

%Before going any further, let's consider how should we create informative training set. Previously work, such as ~\cite{Bychkovsky:2011:LPG}, used a very large dataset(5k image pairs) to tain the regression function.
%However, for practical usage, it's a labor intensive task to adjust too many pairs of examples to train the regression functions. Therefore, it's pretty much desired to have an automatic algorithm
%to select the most representative subset of images for training. The intuition behind this is we want to equally treat all the potential context, and learn hidden CMF for them.

\subsection{Cross-Entropy Based Image Selection}
In example-based photo enhancement, example images that demonstrate a certain enhancement style often need to be manually prepared by human artists. It is a labor intensive task to adjust many images as each image has multiple attributes and regions that can be adjusted. Therefore, it is much desired to pre-select a small number of representative training images to reduce the amount of human work required.
%It is often necessary to try different combinations of enhancement operations on an image before a satisfactory result can be reached.
%Therefore, it is much desired to minimize the number of training images as it would reduce the amount of human effort, which in turn lowers the overall difficulty in obtaining trained photo enhancement models.
On the other hand, to make a learned model achieve a strong prediction capability, it is necessary for the selected training images to have a reasonable coverage of the feature space.

In this section, we introduce a cross-entropy based scheme for selecting a subset of representative training images from a large collection. We first learn a codebook of feature descriptors with $K=400$ codewords by running K-means clustering on feature descriptors collected from all training images. Then every original feature descriptor can find its closest codeword in the codebook via vector quantization, and each image can be viewed as ``a bag of" codewords by quantizing all the feature descriptors in the image. We further build a histogram for every image using the codewords in the codebook as histogram bins. The value in a histogram bin is equal to the number of times the corresponding codeword appears in the image. Let $H^k$ be the histogram for image $\mathcal{I}^k$. For any subset of images $\Omega$ from an initial image collection $\Omega^I$, we compute the accumulated histogram $H^\Omega$ by simply performing elementwise summation over the individual histograms of the images in $\Omega$. We further evaluate the representative power of $\Omega$ using the cross entropy of $H^\Omega$. That is, $\mbox{Entropy}(H^{\Omega})= -\sum_j H^{\Omega}(j) \log H^{\Omega}(j)$, where $H^{\Omega}(j)$ denotes the $j$-th element of $H^\Omega$. A large cross entropy implies that the codewords corresponding to the histogram bins are evenly distributed in the images in $\Omega$ and vice versa.
Thus, to encourage an even coverage of the feature space, the set of selected images essentially need to be the solution of the following expensive combinatorial optimization,
\begin{equation}
\Omega = \arg\max_{\Omega \in \Omega^I} - \sum_j H^\Omega(j) \log H^\Omega(j).
\end{equation}
In practice, we seek an approximate solution by progressively adding one image to $\Omega$ every time until we have a desired number of images in the subset. Every time the added image maximizes the cross entropy of the expanded subset. This process is illustrated in Algorithm 1.

\begin{algorithm} [!t]
\caption{Small Training Set Selection}
\KwIn{A large image collection, $\Omega^I$; The desired number of representative images, $m_d$}
\KwOut{A subset $\Omega$ with $m_d$ images selected from $\Omega^I$}
Initialize $\Omega\leftarrow \emptyset$  \\
\For{ $i=1$ to $m_d$ }
{
   $I^*=\arg\max_{I\in\Omega^I-\Omega} - \sum_j H^{\Omega^\prime}(j) \log H^{\Omega^\prime}(j)$,\\
   \hspace{8mm}where $\Omega^\prime= \Omega \cup \{I\}$;\\
   $\Omega = \Omega \cup \{I^*\}$
}
\end{algorithm}

\section{Overview of Experiments}
Our proposed method is well suited for learning complex and highly nonlinear photo enhancement styles, especially when the style requires challenging spatially varying local enhancements.
%Our proposed method can not only be applied to global color and tone enhancement, but also more challenging spatially varying local enhancements.
%As psychological studies have long confirmed how color can dramatically invoke artistic moods, feelings and even behaviors, interior designers and photographers could adjust their photos heavily in a local manner so as to convey their  very specific and personalized emotion.
%Local enhancement can be performed by segmenting out individual local regions and applying adjustment parameters specifically tailored for them.
Successful local enhancement may not only rely on the content in a specific local region, but also contents in its surrounding areas. In that sense, such operations could easily result in complex effects that require stylistic or even exaggerated color transforms, making previous global methods (e.g., ~\cite{Bychkovsky:2011:LPG}) and local empirical methods (e.g.,~\cite{Kaufman:2012:CAP}) inapplicable. In contrast, our method was designed to address such challenges with the help of powerful contextual features and the strong regression capability of deep neural networks.

To fully evaluate our method, we hired one professional photographer who carefully retouched three different stylistic local effects using hundreds of photos. Section \ref{Label:StylisticEffect} reports experiments we have conducted to evaluate the performance of our method. Although our technique was designed to learn complex local effects, it can be readily applied to global image adjustments without any difficulty. Experiments in Section \ref{Label:GlobalEffect} and the supplemental materials show that our technique achieves superior performance both visually and numerically when compared with other state-of-the-art methods on the MIT-Adobe Fivek dataset. To objectively evaluate the effectiveness of our method, we have further conducted two user studies (Section \ref{sec:user_study}) and obtained very positive results. %Our experiments and user studies have confirmed that the proposed method can not only learn conventional image enhancement styles for better quality (i.e, color and exposure correction), but also learn more generic and stylistic effects.

\subsection{Experimental Setup}
\label{sec:exp_setup}
{\flushleft \textbf{Neural Network Setup}}. Throughout all the experiments in this paper, we use a fixed DNN with one input layer, two hidden layers, and one output layer (Figure \ref{Fig:NN_architecture}). The number of neurons in the hidden layers were set empirically to \textcolor{black}{192}, and the number of neurons in the output layer were set equal to the number of coefficients in the predicted color transform. Our experiments have confirmed that quadratic color transforms can more faithfully reproduce the colors in adjusted images than affine color transforms. Therefore, there are 30 neurons in the output layer, 10 for each of the three color channels.
%We also experimentally found that using more than two hidden layers and more neurons in each hidden layer does not improve the performance on the MIT-Adobe FiveK dataset. This is primarily because a larger number of hidden layers and neurons give rise to an overly large number of parameters that need to be estimated during the training stage.

{\flushleft \textbf{Data Sampling}}. Since we learn pixel-level color mappings, every pixel within the image is a potential training sample. In practice, \textcolor{black}{we segment each image into around 7,000 superpixels, from each of which we randomly select 10 pixels}. Therefore, for example, even if we only have \textcolor{black}{70} example image pairs for learning one specific local effect, \textcolor{black}{the number of training samples can be as large as 4.9 million}. Such a large-scale training set can largely eliminate the risk of overfitting. It typically takes a few hours to finish training the neural network on a medium size training dataset with hundreds of images. Nevertheless, a trained neural network only needs 0.4 second to enhance a 512-pixel wide test image.

{\flushleft \textbf{Image Enhancement with Learned Color Mappings}}.
Once we have learned the parameters (weights) of the neural network, during the image enhancement stage, we apply the same feature extraction pipeline to an input image as in the training stage. That is, we first perform scene parsing and object detection, and then apply graph-based segmentation to obtain superpixels. Likewise, we also extract a feature vector at the centroid of every superpixel, and apply the color transform returned by the neural network to every pixel within the superpixel. Specifically, the adjusted color at pixel $p_i$ is computed as $y_i=\Phi(\Theta,x_{\nu_i})V(c_i)$, where $\nu_i$ is the superpixel that covers $p_i$.

\section{Learning Local Adjustments} \label{Label:StylisticEffect}

\subsection{Three Stylistic Local Effects}\label{sec:msr_effect}

%\begin{figure}[!t]
%\epsfig{file=images/foreground_popout.eps,width=1.0\linewidth,clip=}
%%  \centerline{\includegraphics[width=1.0\linewidth]{images/gallery7.pdf}}
%  \centerline{\hfill Input image \hfill\hfill Our enhanced result \hfill\hfill Ground truth \hfill}
%  \caption{Foreground Pop-Out local effect.}
%  \label{fig:foreground_popout_gallery}
%\end{figure}

\begin{figure*}[!th]
\centering
 \centerline{\includegraphics[width=0.75\linewidth]{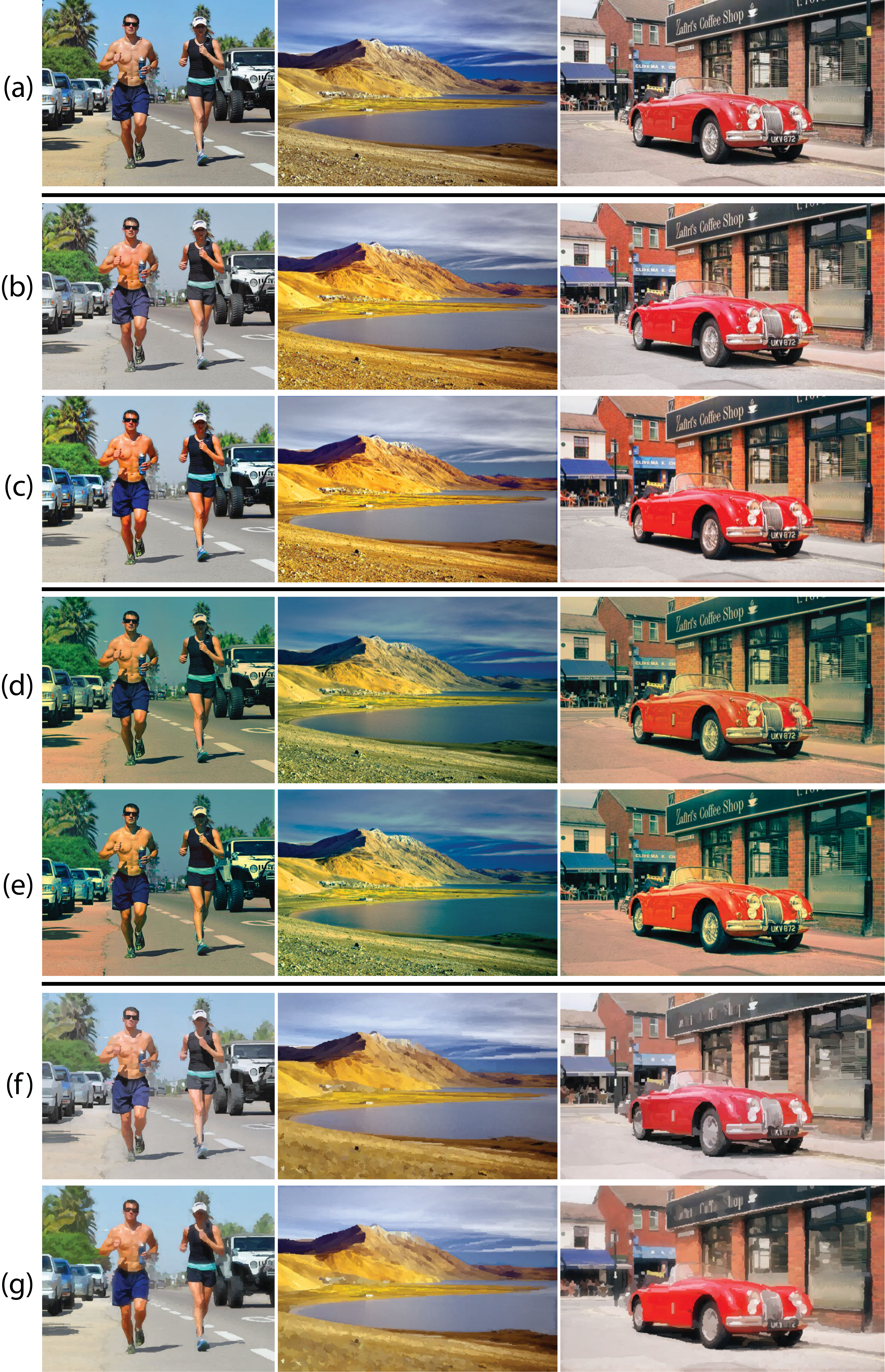}}
%  \centerline{\hfill (a) \hfill\hfill (b) \hfill\hfill (c) \hfill\hfill (d)\hfill\hfill (e) \hfill}
  \caption{Examples of three stylistic local enhancement effects. \textbf{Row (a)}: input images. \textbf{Row (b)\&(c)}: our enhanced results and the groundtruth for the Foreground Pop-Out effect. \textbf{Row (d)\&(e)}: our enhanced results and the groundtruth for the Local Xpro effect. \textbf{Row (f)\&(g)}: our enhanced results and the groundtruth for the Watercolor effect.}
  \label{fig:gallery}
\end{figure*}

\textcolor{black}{We manually downloaded 115 images from \textit{Flickr} and resized them such that their larger dimension has 512 pixels. 70 images were  chosen for training and the remaining 45 images for testing. A professional photographer used Photoshop to retouch these 115 images and produce the datasets for three different stylistic local effects. She could perform a wide range of operations to adjust the images, including selecting local objects/areas with the region selection tool, creating layers with layer masks, blending different layers using various modes, just to name a few. To reduce subjective variation during retouching, she used the ``actions" tool, which records a sequence of operations, which can be repeatedly applied to selected image regions.}

The first local effect \textcolor{black}{"\textbf{Foreground Pop-Out}"} was created by increasing both the contrast and color saturation of \textcolor{black}{foreground salient objects/regions}, while decreasing the color saturation of the background. Before performing these operations, foreground salient regions need to be interactively segmented out using region selection tools in Photoshop. \textcolor{black}{Such segmented regions were only used for dataset production, and they are not used in our enhancement pipeline.} This local effect makes foreground objects more visually vivid while making the background less distractive. Figure \ref{fig:gallery} \textcolor{black}{(b) and (c) show three examples of our automatically enhanced results and groundtruth results from the photographer}.
%The photographer ran our image selection algorithm to choose 50 training photos from the MIT-Adobe FiveK dataset, and manually adjusted these 50 photos to enhance the first effect. This process produced 50 image pairs, which were used as our training dataset. We also randomly chose 50 different photos from other sources as the testing dataset.
Refer to the supplemental materials for the training data as well as our enhanced testing photos.
%Figure \ref{fig:Compare_context_first} shows one such test example. Note that the enhanced image predicted by our approach is almost the same as the one produced by the photographer.

Our second effect \textcolor{black}{"\textbf{Local Xpro}"} was created by generalizing the popular "cross processing" effect in a local manner. \textcolor{black}{Within Photoshop, the photographer first predefined multiple "Profiles", each of which is specifically tailored for one of the semantic categories used in scene parsing and object detection in section \ref{sec:context_ftr}}.
\textcolor{black}{All the profiles share a common series of operations, such as the adjustment of individual color channels, color blending across color channels, hue/saturation adjustment as well as brightness/contrast manipulation, just to name a few. Nonetheless, each profile defines a distinct set of adjustment parameters tailored for its corresponding category. When retouching a photo, the photographer used region selection tools to isolate image regions and then applied one suitable profile to each image region according to the specific semantic content within that region. To avoid artifacts along region boundaries, she could also slightly adjust the color/tone of local regions after the application of profiles.} Although the profiles roughly follow the "cross processing" style, the choice of local profiles and additional minor image editing were heavily influenced by the photographer's personal taste which can be naturally learned through exemplars.
%In practice, as different photographers might have their own unique styles to convey different mood and feelings, in order to digitize those  personalized effects, one must use exemplar-based methods to do reverse-engineer, any empirical and heuristic method would not easily work.
%For this effect, we collected 90 photos and asked the photographer to retouch each of them. We then used 60 photos for training and the remaining 30 for testing.
Figure \ref{fig:gallery} (d)\&(e) \textcolor{black}{show three examples in this effect, and compare our enhanced results against groundtruth results.}
Figure \ref{fig:teaser} shows another example of this effect.

To further increase diversity and complexity, we asked the photographer to create a third local effect \textcolor{black}{"\textbf{Watercolor}"}, which tries to mimic certain aspects of the "watercolor" painting style. For example, watercolors tend to be brighter with lower saturation. Within a single brush region, the color variation also tends to be limited.
%This is because, when people draw the brush, the colors for all the pixels underneath that for that brush is supposed to be consistently the same.
\textcolor{black}{The photographer first applied similar operations as in the Foreground Pop-Out effect to the input images, including increasing both contrast and saturation of foreground regions as well as decreasing those of background regions. In addition, the brightness of both foreground and background regions are increased by different amounts.} %This brought out the foreground/salient objects while deemphasizing the background.
She further created two layers of brush effects from the same brightened image, using larger ``brushes" on one layer and a smaller one on the other. On the first layer, the brush size for the foreground and the background are also different. Finally, these two layers are composited together using the 'Lighten' mode in Photoshop. Overall, this effect results in highly complex and spatially varying color transforms, which force the neural network to heavily rely on local contextual features during regression.

\begin{figure}[t]
\centerline{\includegraphics[width=1.0\linewidth]{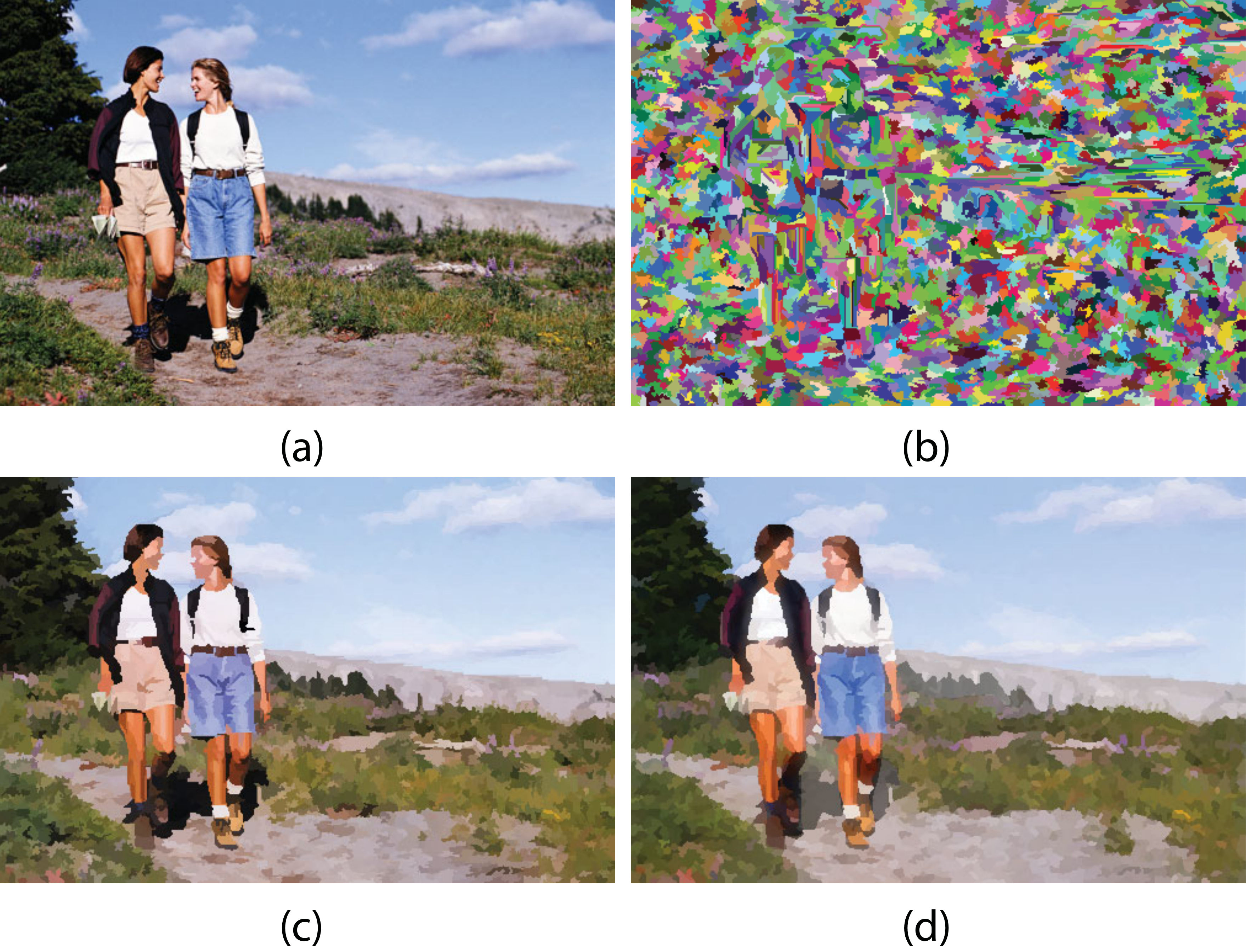}}
\caption{An example of Watercolor local effect. \textbf{(a)}: input image. \textbf{(b)}: a visualization of superpixels used for simulating brush strokes. Each superpixel is filled with a random color. \textbf{(c)}: our enhanced result. \textbf{(d)}: the ground truth. }
\label{fig:watercolor_seg_vis}
\end{figure}

%In our experiments, we randomly chose 70 photos for training and the rest for testing.
Figure \ref{fig:gallery} (f)\&(g) \textcolor{black}{show the enhanced results of three testing examples and their corresponding groundtruth results}. \textcolor{black}{To simulate brush strokes, after applying the same color transform to all pixels in a superpixel, we calculate the average color within the superpixel and fill the superpixel with it. See another example of Watercolor effect as well as visualized superpixels in Fig \ref{fig:watercolor_seg_vis}.} Our automatic results look visually similar to the ones produced by the photographer. Refer to the supplemental materials for more examples enhanced with this effect. Note that our intention here is not rigorously simulating watercolors, but experimentally validating that our technique is able to accurately learn such complex local adjustments.

\textcolor{black}{To successfully learn an enhancement effect, it is important to make the adjustments on individual training images consistent. In practice, we have found the following strategies are helpful in increasing such consistency across an image set. First, as artistic adjustment of an image involves the personal taste of the photographer, the result could be quite different from different photographers. Therefore, we always define a retouching style using photos adjusted by the same photographer. That means, even for the same input content, retouched results by different photographers are always defined as different styles. Second, we inform the photographer the semantic object categories that our scene parsing and object detection algorithms are aware of. Consequently, she can apply similar adjustments to visual objects in the same semantic category. Third, we use the "actions" tool in Photoshop to faithfully record the "Profiles" that should be applied to different semantic categories. This improves the consistency of color transforms applied to image regions with similar content and context.
}

%\begin{figure}[!th]
%\centering
%\epsfig{file=img_colormapping/color_mapping.eps,width=0.8\linewidth,clip=}
%\caption{Color mapping scatter plots. \textbf{On the left (from top to bottom)}, input image, semantic label map and the groundtruth for the Local Xpro effect. \textbf{On the right}, color mapping scatter plots for semantic regions. Each semantic region has three scatter plots corresponding to its $L$, $a$, $b$ color channels. \hl{Each scatter plot visualized two sets of points, which take the original value of a channel as the horizontal coordinate, and respectively the predicted (red) and groundtruth (blue) adjusted values of that channel as the vertical coordinate.}}
%\label{fig:color_mapping}
%\end{figure}

\subsection{Spatially Varying Color Mappings}

\begin{figure}[!h]
  \centerline{\includegraphics[width=1.0\linewidth]{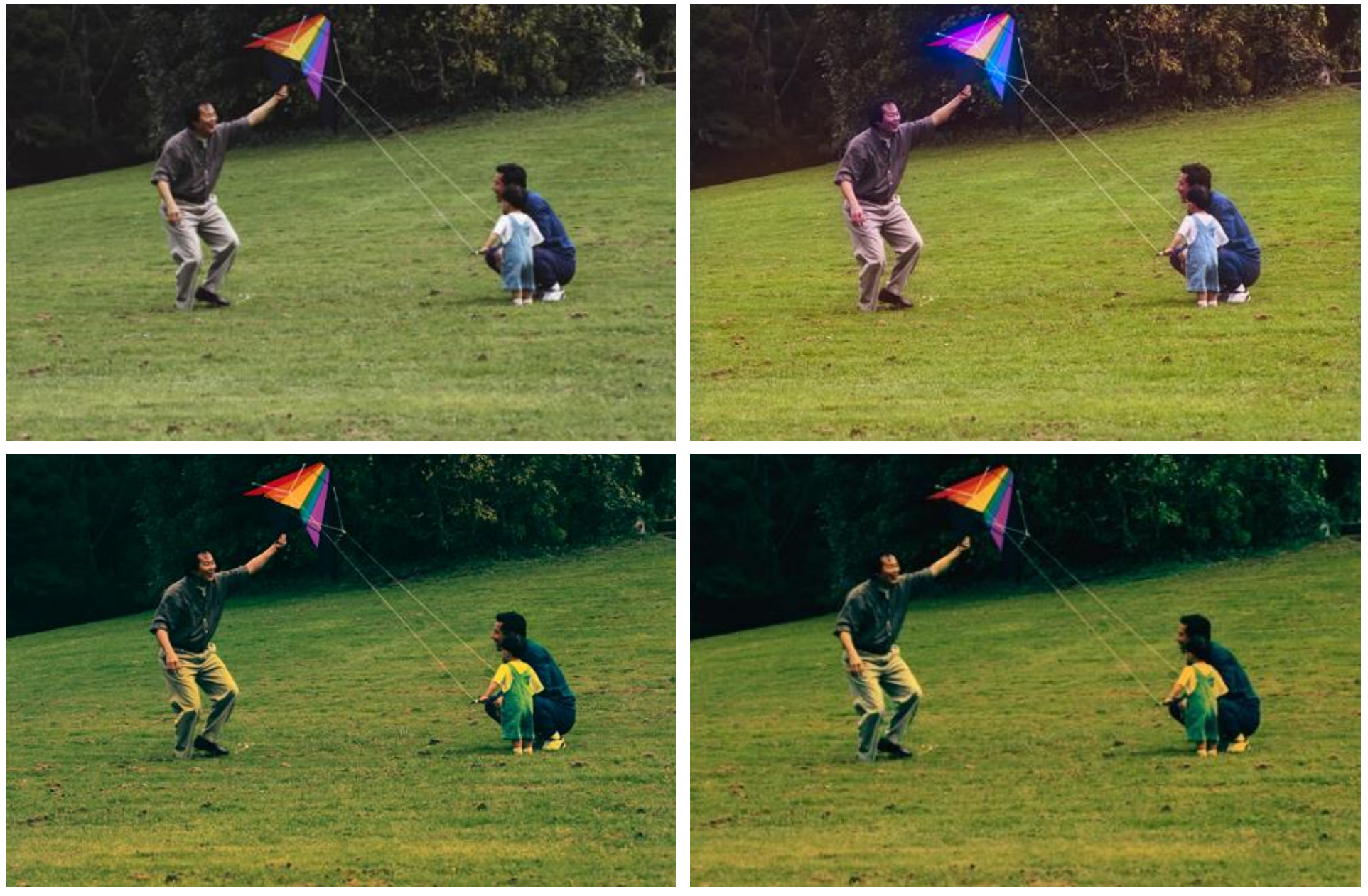}}
  \caption{Comparison with ~\protect\cite{Wang:2011:EIC} on the Local Xpro effect. \textbf{Top Left}: Input image; \textbf{Top Right}: enhanced image by ~\protect\cite{Wang:2011:EIC}; \textbf{Bottom Left:} enhanced image by our approach; \textbf{Bottom Right:} enhanced image by photographer. The enhanced image by our approach is closer to the ground truth generated by the photographer.}\label{fig:Compare_wang}
\end{figure}

\begin{figure*}[!th]
\centering
\centerline{\includegraphics[width=1.0\linewidth]{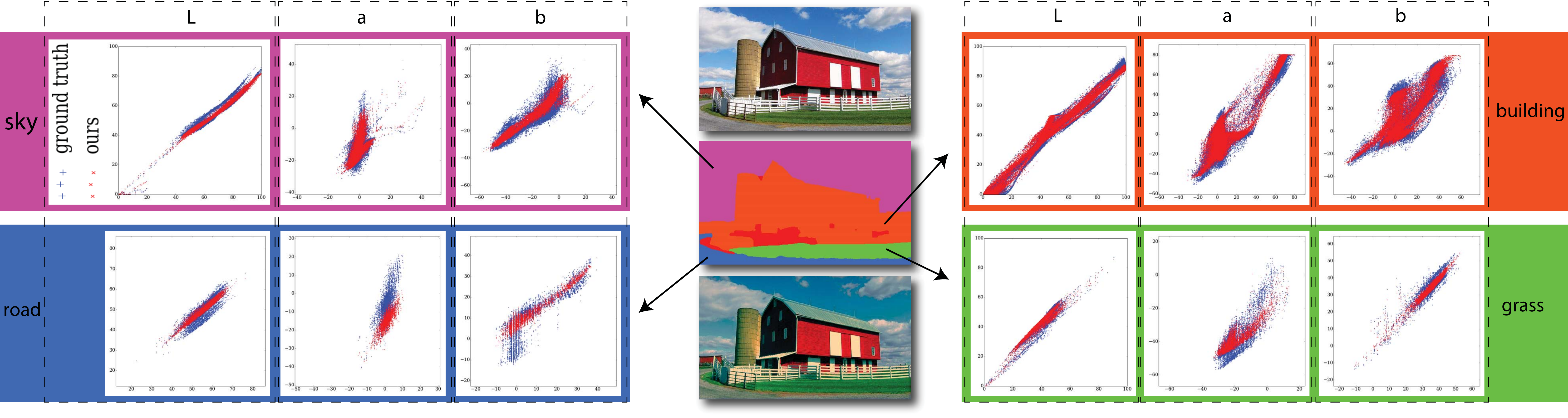}}

\caption{Scatter plots of color mappings. \textbf{Middle (from top to bottom)}: input image, semantic label map and the groundtruth for the Local Xpro effect. \textbf{Left and right}: color mapping scatter plots for four semantic regions. Each semantic region has three scatter plots corresponding to its $L$, $a$, $b$ color channels. Each scatter plot visualizes two sets of points, which take the original value of a channel as the horizontal coordinate, and respectively the predicted (red) and groundtruth (blue) adjusted values of that channel as the vertical coordinate.}
\label{fig:color_mapping}
\end{figure*}

\textcolor{black}{It is important to point out that the underlying color mappings in the local effect datasets are truly not global. They spatially vary within the image domain. To verify this, we collect pixels from each semantic region of an image. By drawing scatter plots for different semantic regions using pixel color pairs from the input and retouched images, we are able to visualize the spatially varying color transforms. See such an example in Figure \ref{fig:color_mapping}, which clearly shows that the color transforms differ in the sky, building, grass and road regions. Also, we can see that our method can successfully learn such spatially varying complex color transforms.}
We further conducted a comparison against \cite{Wang:2011:EIC}, which adopts a local piecewise approximation approach. However, due to the lack of discriminative contextual features, their learned adjustment parameters tend to be similar across different regions (Figure \ref{fig:Compare_wang}).
 %Figure \ref{} shows another side-by-side comparison with the method in \cite{Wang:2011:EIC}. Although they applied local piecewise approximation, due to the lack of discriminative contextual features, adjustment parameters for different regions tend to be averaged.Refer to the supplemental materials for additional results.

%\begin{table}[h]
%\centering
%\begin{tabular}{@{}cccc@{}}
%\toprule
%\multicolumn{1}{l}{\textit{}} & \multicolumn{1}{l}{No context} & \multicolumn{1}{l}{Naive context} & \multicolumn{1}{l}{Our context} \\ \midrule
%Xpro Effect                   & 15.7                           & 11.9?                              & \textbf{10.7}                   \\
%\midrule
%Painting Effect               &  12.3                              & 8.5                                & \textbf{6.1}                    \\ \bottomrule
%\end{tabular}
%\caption{Mean L2 distance in CIElab space for comparisons of cases: without context feature, with the naive context feature and with our proposed contextual feature respectively}
%\label{tab:Context}
%\end{table}

\subsection{Generalization Capability}
\begin{figure}[!t]
 \centerline{\includegraphics[width=1.0\linewidth]{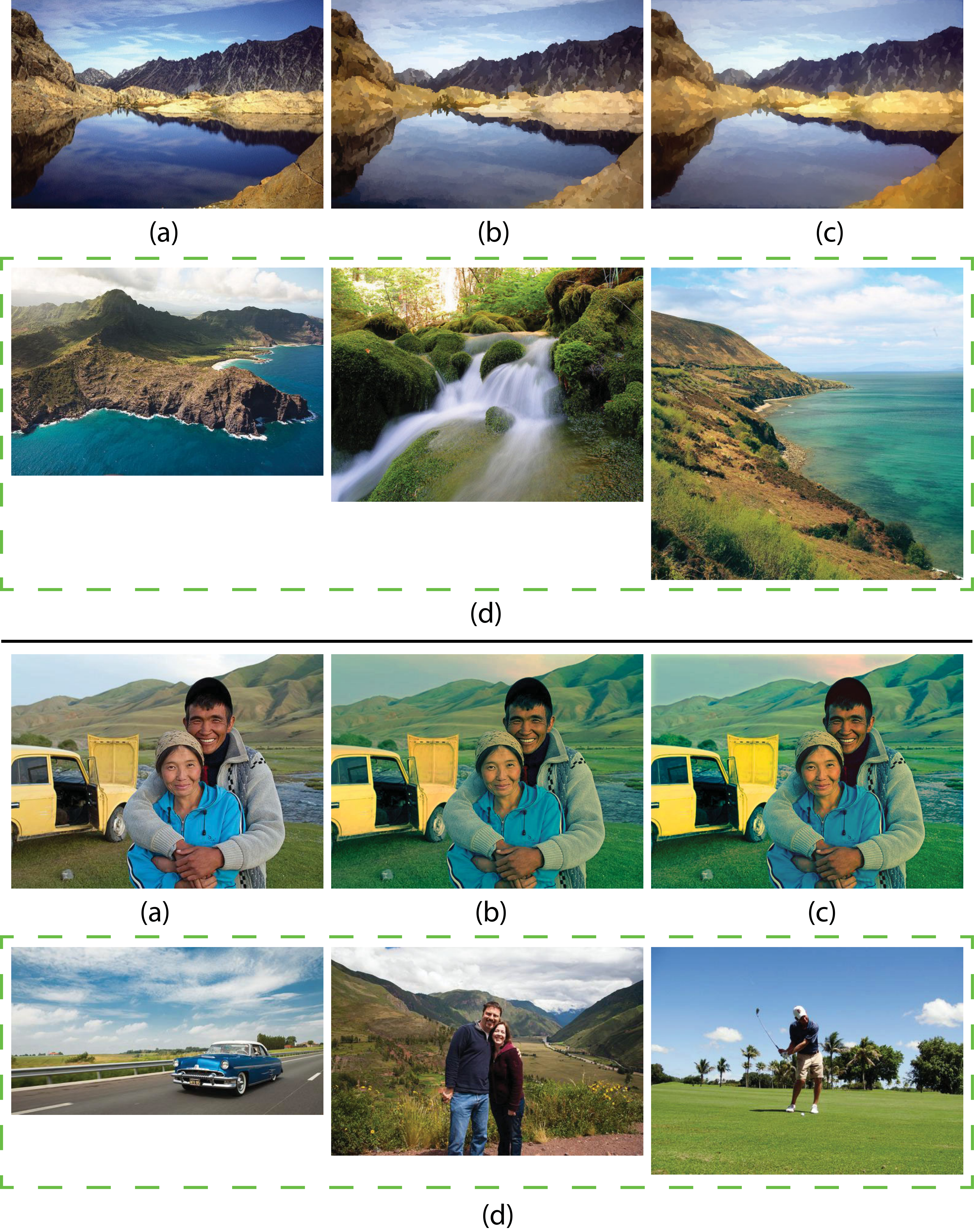}}
%  \centerline{\hfill Input image \hfill\hfill Our result \hfill\hfill Ground truth \hfill}
  \caption{Two examples of novel image enhancement. \textbf{Top}: an example of the Watercolor effect. \textbf{Bottom}: an example of the Local Xpro effect. In each example, \textbf{(a)}: input image, \textbf{(b)}: our enhanced result, \textbf{(c)}: ground truth, \textbf{(d)}: training images most similar to the input image. Note that the input images in these examples have significant visual differences from any images in the training set.}
  \label{fig:novel_test}
\end{figure}

Here we verify the generalization capability of the DNN based photo adjustment models we trained using $70$ image pairs. As mentioned earlier, the actual number of training samples far exceeds the number of training image pairs because we use thousands of superpixels within each training image pair. As shown in Fig. \ref{fig:novel_test}, we apply our trained models to novel testing images with significant visual differences from any images in the training set.
The visual objects in these images have either unique appearances or unique spatial configurations. To illustrate this, we show the most similar training images, which not only share the largest number of object and region categories with the testing image, but also have a content layout as similar as possible.
In Fig \ref{fig:novel_test} top, the mountain in the input image has an appearance and spatial layout that are different from the training images. In Fig \ref{fig:novel_test} bottom, the appearances and spatial configuration of the car and people are also quite different from those of the training images. In despite of these differences, our trained DNN models are still able to adjust the input images in a plausible way.

\subsection{Effectiveness of Contextual Features}
We demonstrate the importance of contextual features in learning local adjustments in this subsection.
%we simply exclude contextual features and only use pixelwise features and global features when learning color mappings.
\textcolor{black}{First, we calculate the $L^2$ distance in the 3D CIELab color space between input images and ground truth produced by the photographer for all local effect datasets as shown in the second column of Table \ref{tb:local_dataset}. They numerically reflect the magnitude of adjustments the photographer made to the input images. Second, we numerically compare the testing errors of our enhanced results with and without the contextual feature in the third and fourth columns of Table \ref{tb:local_dataset}}. Our experiments show that without contextual features, testing errors of our enhanced results tend to be relatively high. \textcolor{black}{The mean $L^2$ error in the 3D CIELab color space reaches {\bf 9.27}, {\bf 9.51} and {\bf 9.61} respectively for the Foreground Pop-Out, Local Xpro and Watercolor effects}. On the other hand, by including our proposed contextual feature, all errors drop significantly to {\bf 7.08}, {\bf 7.43} and {\bf 7.20}, indicating the necessity of such features.

\begin{figure*}[!th]
  \centerline{\includegraphics[width=1.0\linewidth]{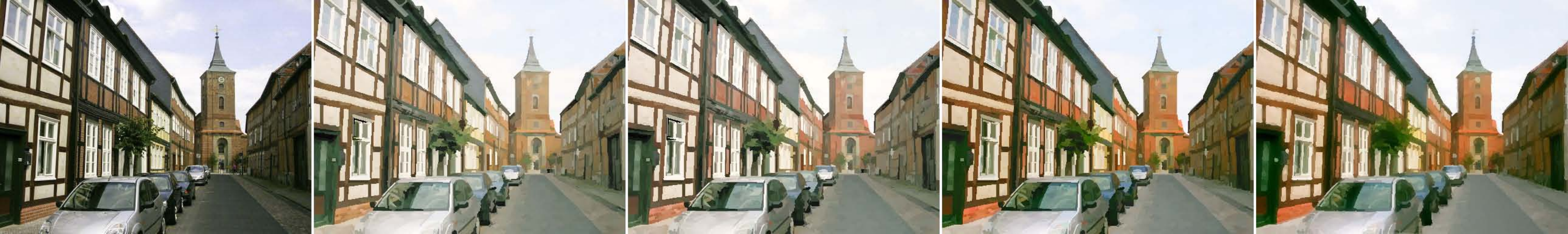}}
  \centerline{\hfill Input Image \hfill\hfill Without Context \hfill\hfill Simple Context \hfill\hfill Our Context \hfill\hfill Ground Truth \hfill}
  \caption{Effectiveness of our contextual features. \textbf{Left:} Input; \textbf{Middle Left:} enhanced without context; \textbf{Middle:} enhanced with simple contextual features (from a single pooling region); \textbf{Middle Right:} enhanced with our contextual features; \textbf{Right:} ground truth. It is obvious that among all enhanced results, the one enhanced using our contextual features is the closest to the ground truth.}
  \label{fig:Compare_context}
\end{figure*}

\begin{table}[hb]
\tbl{Statistics of three local effects and the mean $L^2$ testing errors. TE=Testing Error.}{
%\begin{center}
	\begin{tabular}{|p{2.4cm}|p{1.3cm}|p{1.2cm}|p{1.2cm}|}
	\hline
	 Effect & ground truth $L^2$ distance & TE w/o context & TE w/ context \\ \hline \hline
	Foreground Pop-Out & 13.86 & 9.27 & \textbf{7.08}\\ \hline
	Local Xpro & 19.71 & 9.51 & \textbf{7.43}  \\ \hline
	Watercolor & 15.30 & 9.61 & \textbf{7.20} \\ \hline
	\end{tabular}
%\end{center}
}
\label{tb:local_dataset}
\end{table}

To validate the effectiveness of our multiscale spatial pooling schema in our contextual feature design, we have experimented with a simpler yet more intuitive contextual feature descriptor with just one pooling region (vs. our 28 multiscale regions) at the same size as our largest region, and found that such simple contextual features are helpful in reducing the errors but not as effective as ours. \textcolor{black}{Taking the local Watercolor painting effect as an example, we observed the corresponding mean $L^2$ error is {\bf 8.28}, which drops from {\bf 9.61}, but still obviously higher than our multiscale features {\bf 7.20}}. This is because, with multiscale pooling regions, our features can achieve a certain degree of translation and rotation invariance, which is crucial for the histogram based representation. We have also performed visual comparisons. Fig. \ref{fig:Compare_context} shows one such example. We can see that without our contextual feature, local regions in the enhanced photo might exhibit severe color deviation from the ground truth.

\subsection{Effectiveness of Learning Color Transforms}
\textcolor{black}{
As shown in Figure \ref{fig:vis_color_transform}, the use of color transforms helps absorb high-frequency color variations and enables DNN to regress the spatially smooth but otherwise highly nonlinear part of the color mapping. To highlight the benefits of using color transforms, we train a different DNN to regress the retouched colors directly. The DNN architecture is similar to the one described in section \ref{sec:exp_setup} except that there are only 3 neurons in the output layer, which represent the enhanced CIELab color. We compare the testing $L^2$ errors on the Foregronud Pop-Out and Local Xpro datasets in Table \ref{tb:comp_regress_target}. On both datasets, the testing error increases by more than $20\%$ which indicates the use of color transforms is beneficial in our task.
}

\begin{table}[hb]
\tbl{Comparison of $L^2$ testing errors obtained from deep neural networks using and without using quadratic color transforms.}{
%\begin{center}
	\begin{tabular}{|p{2.4cm}|p{2.0cm}|p{2.cm}|}
	\hline
	 Effect & w/o transform & w/ transform \\ \hline \hline
	Foreground Pop-Out & 8.90 & \textbf{7.08}\\ \hline
	Local Xpro & 9.01 & \textbf{7.43}  \\ \hline
	\end{tabular}
%\end{center}
}
\label{tb:comp_regress_target}
\end{table}

\subsection{DNN Architecture}
\textcolor{black}{The complexity of our DNN based model is primarily determined by the number of hidden layers and the number of neurons in each layer. Note that the complexity of the DNN architecture should meet the inherent complexity of the learning task. If the DNN did not have the sufficient complexity to handle the given task, the trained model would not even be able to accurately learn all the samples in the training set. On the other hand, if the complexity of the DNN exceeds the inherent complexity of the given task, there exists the risk of overfitting and the trained model would not be able to generalize well on novel testing data even though it could make the training error very small.}

\begin{figure}[!t]
 \centerline{\includegraphics[width=1.0\linewidth]{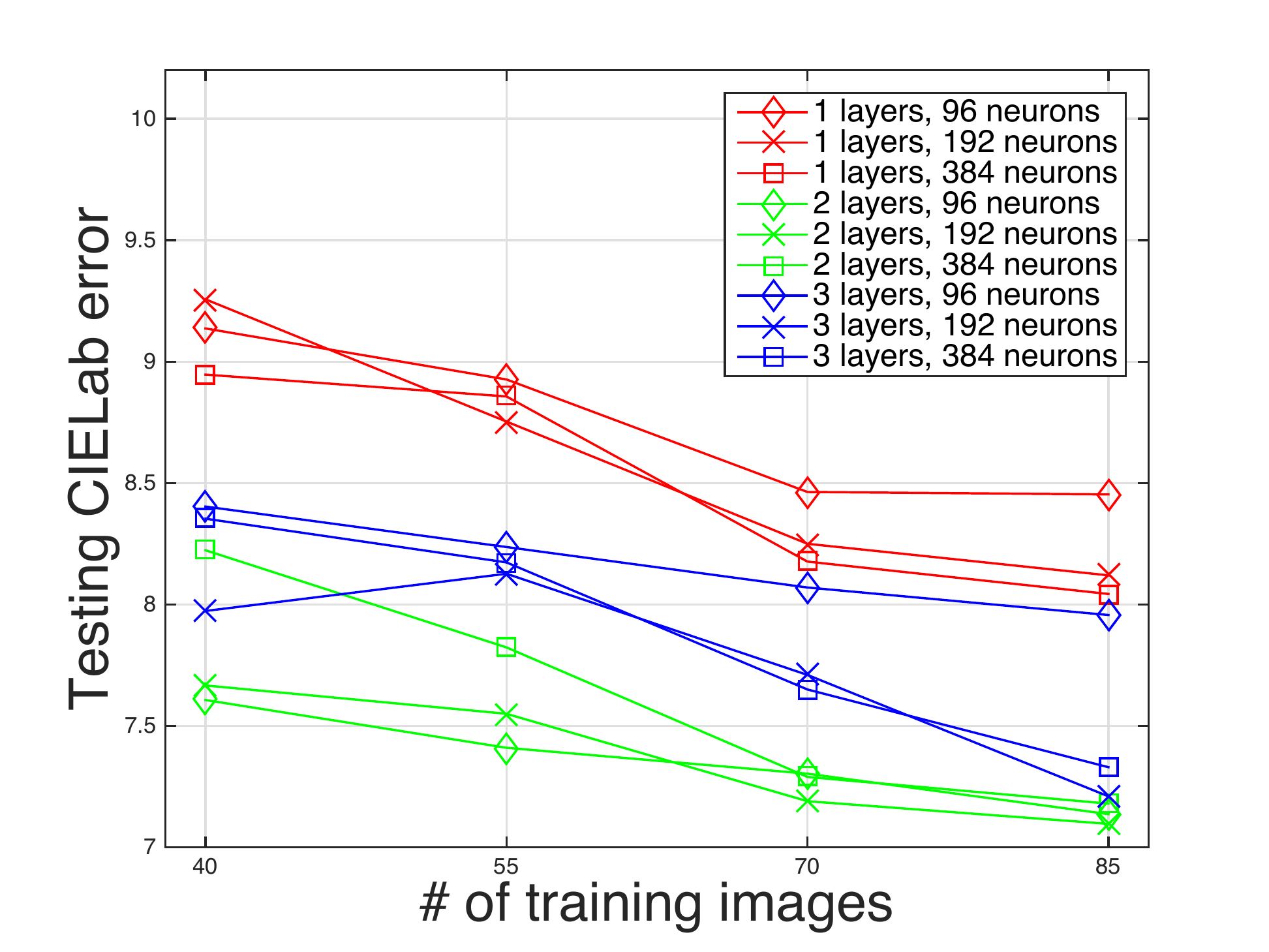}}
  %\centerline{\hfill Input image \hfill\hfill Semantic label map \hfill\hfill Our result \hfill\hfill Ground truth \hfill}
  \caption{ Testing error vs number of training images for DNNs of various architectures. Error bars are omitted for clarity.}
  \label{fig:dnn_design}
\end{figure}

\textcolor{black}{The nature of the learning task in this paper is a regression problem. It has been shown that a feedforward neural network with a single hidden layer~\cite{Hornik:1989:MFN} can be used as a universal regressor and the necessary number of neurons in the hidden layer varies with the inherent complexity of the given regression problem. In practice, however, it is easier to achieve a small training error with a deeper network that has a relatively small number of neurons in the hidden layers. To assess the impact of the design choices of the DNN architecture, we evaluate DNNs with a varying number of hidden layers and neurons. We keep a held-out set of 30 images for validation and vary the number of training images from $40$ to $85$ at a step size of $15$ to evaluate the impact of the size of the training set. We repeat the experiments for five times with random training and testing partitions and report the averaged results. The Foreground Pop-Out dataset is used in this study. Fig \ref{fig:dnn_design} summarizes our experimental results. Overall, neural networks with a single hidden layer deliver inferior performance than deeper networks. DNNs with 3 hidden layers do not perform as well as those with 2 hidden layers. For a DNN with 2 hidden layers, when the number of training images exceeds 70, the testing error does not significantly improve any more. In summary, DNNs with 2 hidden layers achieve low testing errors and execute faster than those with 3 hidden layers in both training and testing stages. Therefore, we finally use a DNN with $2$ hidden layers and $192$ neurons each throughout this paper.
}

\subsection{Comparison with Other Regression Methods}

\begin{table}[h]
\tbl{Comparison of $L^2$ testing errors obtained from different regressors. }{
%\begin{center}
	\begin{tabular}{|p{2.4cm}|p{1.2cm}|p{1.2cm}|p{1.2cm}|}
	\hline
	 Effect & Lasso & Random Forest & DNN \\ \hline \hline
	Foreground Pop-Out & 11.44 & 9.05 & \textbf{8.90}\\ \hline
	Local Xpro & 12.01 & \textbf{7.51} & 9.01  \\ \hline
	Watercolor & 9.34 & 11.41 & \textbf{9.22} \\ \hline
	\end{tabular}
%\end{center}
}
\label{tb:comp_regressor}
\end{table}

\begin{figure*}[!th]
\centerline{\includegraphics[width=1.0\linewidth]{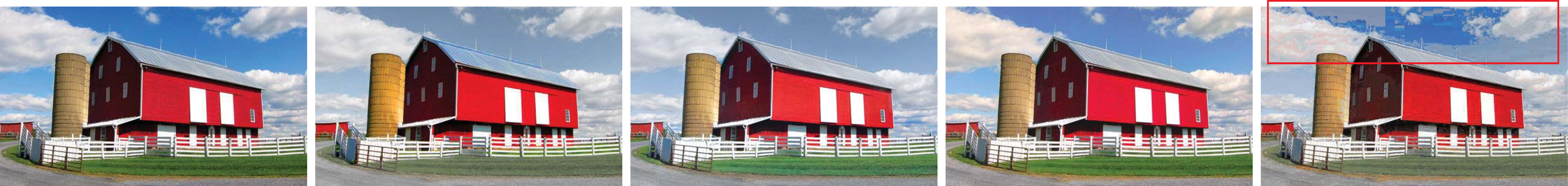}}

   \centerline{\hfill Input Image \hfill\hfill Ground truth  \hfill\hfill Our result  \hfill\hfill Lasso \hfill\hfill Random forest  \hfill}
   \caption{Visual comparison against Lasso and a random forest. Note an area with blocky artifacts in the result of the random forest is highlighted. }
\label{fig:vis_comp_regressor}
\end{figure*}

Our DNN proves to be effective for regressing spatially varying complex color transforms on the three local effect datasets. It is also of great interest to evaluate the performance of other regressors on our datasets. Specifically, we chose to compare DNN against two popular regression methods, Lasso~\cite{tibshirani1996regression} and random forest~\cite{breiman2001random}.
%On one hand, the cost function of Lasso \cite{tibshirani1996regression} includes both a squared error term and a $L1$ penalty term on weights. It tends to give rise to sparse solutions which can generalize well to unseen data. On the other hand, a random forest \cite{breiman2001random} is an ensemble learning model which uses the bagging technique to obtain different regression trees and selects a random subset of feature at each tree node splitting. The final prediction is the averaged result of all trees.
Both Lasso and random forest are scalable to the large number of training samples used in DNN training.
%Due to the formulation, neither Lasso or random forest can regress the color transform if $L2$ error of pixel color is used as the cost function. Therefore,
We use Lasso and a random forest to directly regress target CIELab colors using the same feature set as in DNN training, including pixelwise features, global features and contextual features. The hyperparameters of both Lasso and the random forest are tuned using cross validation.
To make a fair comparison, our DNN is also adapted to directly regress the target CIELab colors. A comparison of $L^2$ errors is summarized in Table \ref{tb:comp_regressor}. The DNN significantly outperforms Lasso on the Foreground Pop-Out and Local Xpro datasets, and obtains slightly lower errors on the Watercolor dataset. Compared with the random forest, the DNN obtains lower testing errors on both Foreground Pop-Out and Watercolor datasets. On the Local Xpro dataset, the random forest obtains lower numerical errors than that of the DNN. However, after visual inspection, we found that colors generated by the random forest are not spatially smooth and blocky artifacts are prevalent in the enhanced images, as shown in Figure \ref{fig:vis_comp_regressor}. This is because regression results from a random forest are based on values retrieved from various leaf nodes, and spatial smoothness of these retrieved values cannot be guaranteed. In contrast, our trained DNN generates spatially smooth colors and does not give rise to such visual artifacts.

%A comparison of $L^2$ errors is summarized in Table \ref{tb:comp_regressor}. DNN significantly outperforms Lasso in all three local effect datasets, and obtains lower testing errors than the random forest on both Foreground Pop-Out and Watercolor datasets. On the Local Xpro dataset, the random forest slightly outperforms DNN. However, after visual inspection, we found that the color mappings generated by the random forest are not spatially smooth and blocky artifacts are prevalent in the enhanced images, as shown in Figure \ref{fig:vis_comp_regressor}.}

\section{Learning Global Adjustments}\label{Label:GlobalEffect}

\begin{table}[!b]
\tabcolsep10pt
\tbl{Comparison of mean $L^2$ errors obtained with our method and previous methods on the MIT-Adobe FiveK dataset. The target style is \textit{Expert C}.}{
%\begin{center}
	\begin{tabular}{|p{1.6cm}|c|c|c|}
	\hline
	Method & 2500(L) & Ran. 250(L,a,b) & H.50(L,a,b)\\ \hline \hline
	~\cite{Bychkovsky:2011:LPG} & 5.82 &N/A & N/A\\ \hline
	~\cite{Hwang:2012:CAL} & N/A & 15.01 & 12.03  \\ \hline
	Our method & \textbf{5.68} & \textbf{9.85}& \textbf{8.36}  \\ \hline
	\end{tabular}
%\end{center}
}
\label{tb:dataset}
\end{table}

\subsection{MIT-Adobe FiveK Dataset}\label{sec:fivek}

The MIT-Adobe FiveK dataset ~\cite{Bychkovsky:2011:LPG} contains 5000 raw images, each of which was retouched by five well trained photographers, which results in five groups of global adjustment styles. As we learn pixel-level color mappings, there would be 175 million of training samples in total if half of the images are used for training.
%In our method, each image was first divided into 10K superpixels and seven pixels were selected within each superpixel. Since we learn pixel-level color mappings, every selected pixel is a training sample. This results in 175 million training samples in total if half of the images are used for training.
%In the training stage, we randomly partition the training set into 300 batches, each of which is further divided into a series of mini-batches (128 samples per min-batch), and then perform stochastic gradient descent on mini-batches sequentially. It typically takes 21 hours to finish 25 epoches before convergence. Although training is slow, a trained neural network only needs 0.4 second to enhance a 500-pixel wide image.

\begin{figure}[!t]
  \centerline{ \includegraphics[width=1.0\linewidth]{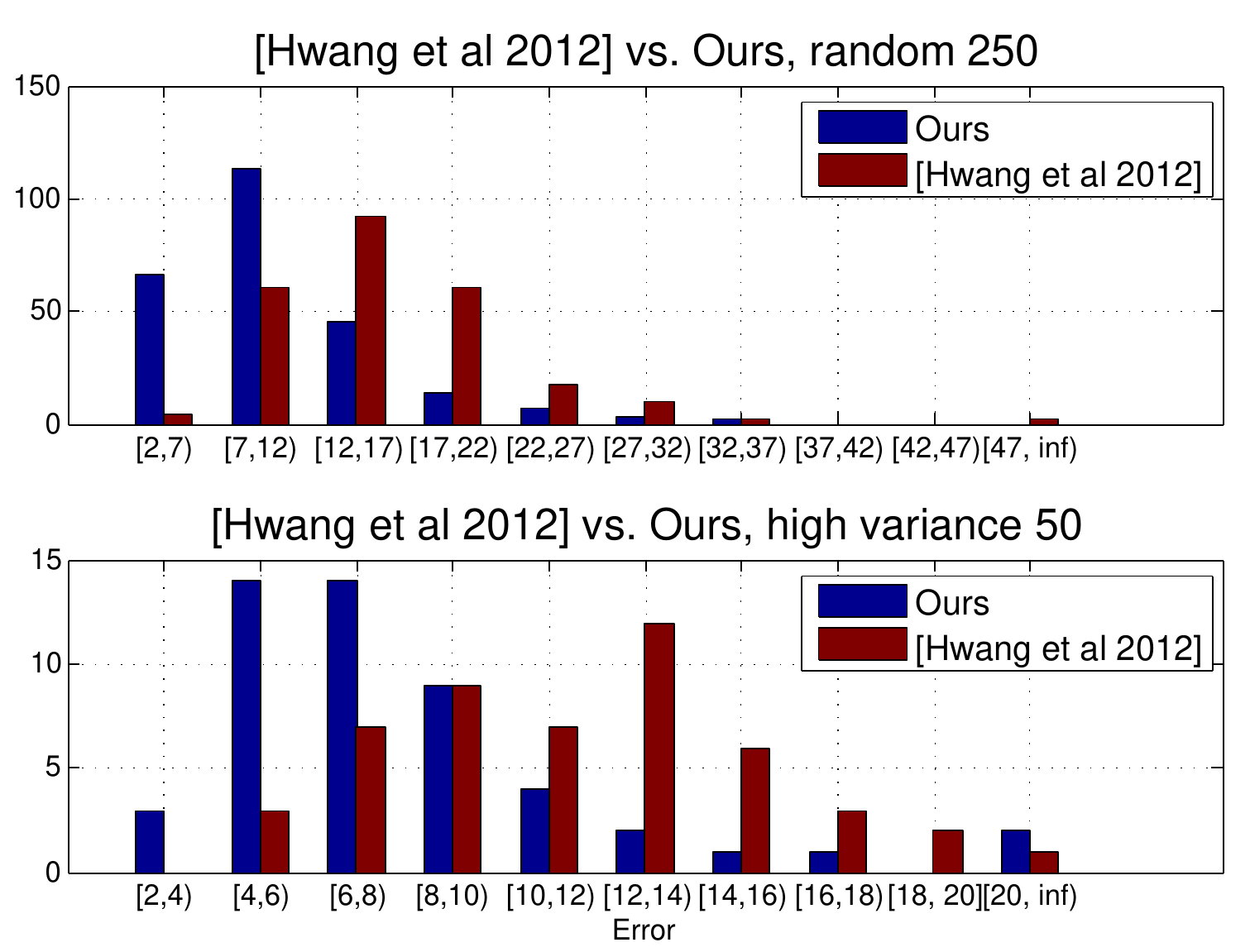}}

   \caption{$L^2$ error distributions. Note that our method produces smaller errors on both testing datasets.}
\label{fig:error_hostogram}
\end{figure}

\begin{figure*}[!th]
   \centerline{\includegraphics[width=1.0\linewidth]{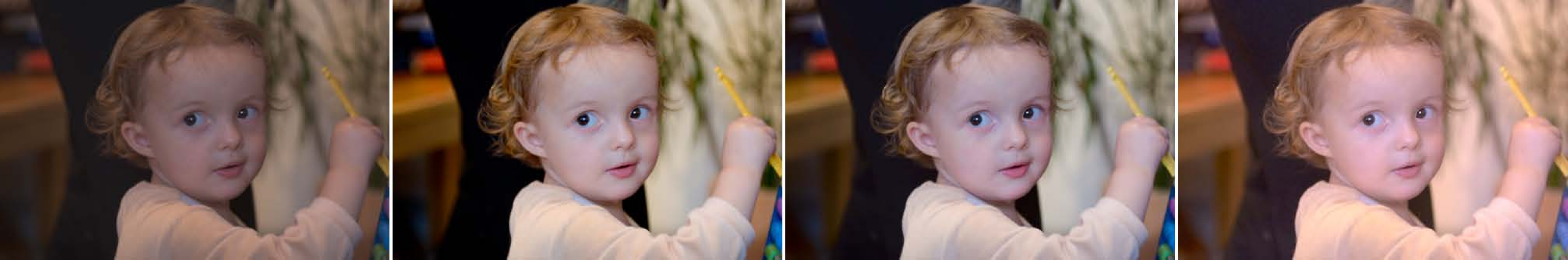}}
   \centerline{\hfill Input Image \hfill\hfill Ground Truth \hfill\hfill Our Result \hfill\hfill [Hwang {\em et al.} 2012] \hfill}
   \caption{Visual comparison with ~{\protect\cite{Hwang:2012:CAL}}. \textbf{Left:} Input image; \textbf{Middle Left:} groundtruth enhanced image by expert C; \textbf{Middle Right:} enhanced image by our approach; \textbf{Right:} enhanced image by ~{\protect\cite{Hwang:2012:CAL}}.}
\label{fig:eccv_compare}
\end{figure*}

We have compared our method with \cite{Hwang:2012:CAL} using the same experimental settings and testing datasets in that work. Two testing datasets were used in \cite{Hwang:2012:CAL}. (1)``Random 250": 250 randomly selected testing images from group C of the MIT-Adobe FiveK dataset (hence 4750 training images) and (2) ``High Variance 50": 50 images selected for testing from group C of the MIT-Adobe FiveK dataset (hence 4950 images for training). Comparison results on numerical errors are shown in the second and third columns of Table \ref{tb:dataset}, from which we can see our method is capable of achieving much better prediction performance in terms of mean $L^2$ errors on both predefined datasets. Figure \ref{fig:error_hostogram} further shows the error histograms of our method and \cite{Hwang:2012:CAL} on these two testing datasets. The errors produced by our method are mostly concentrated at the lower end of the histograms.
Figure \ref{fig:eccv_compare} shows a visual comparison, from which we can see our enhanced result is closer to the ground truth produced by the photographer.
\textcolor{black}{
% Our approach outperforms the matching-based approach in \cite{Hwang:2012:CAL}, as confirmed in Table \ref{tb:dataset} and Figs. \ref{fig:eccv_compare}, \ref{fig:error_hostogram} and \ref{fig:user_study}.
Such performance differences could be explained as follows.
The technique in \cite{Hwang:2012:CAL} is based on nearest-neighbor search, which requires a fairly large training set that is slow to search. As a result, this technique divides similarity based search into two levels. It first searches for the most similar images and then the most similar pixels within them. While this two-level strategy accelerates the search, a large percentage of similar pixels does not even have the chance to be utilized because the search at the image level leaves out dissimilar images that may still contain many similar pixels. On the other hand, our deep neural network based method is a powerful nonlinear regression technique that considers all the training data simultaneously. Thus our method has a stronger extrapolation capability than the nearest-neighbor based approach in \cite{Hwang:2012:CAL}, which only exploits a limited number of nearest neighbors. For the same reason, the nearest-neighbor based approach in \cite{Hwang:2012:CAL} is also more sensitive to noisy and inconsistent adjustments in the training data.
}
In another comparison with \cite{Bychkovsky:2011:LPG}, we follow the same setting used in that work, which experimented on 2500 training images from group C and reported the mean error on the L channel (CIELAB color space) only.
As shown in the first column of Table \ref{tb:dataset}, we obtained a slightly smaller mean error on the L channel on the remaining 2500 testing images.

%{\color{red} Add comparisons of error histograms on both ``Random 250" and ``High Variance 50".}
%'Random 250': the range of errors (x-axis of the histogram) is [2,48]
%            Our error histogram       	[66, 113, 45, 14,   7,   3,   2,  0,  0,  0 ]
%            Hwang's error histogram 	[  4,    61, 92, 61, 18, 10,  2,  0,  0,  2 ]
%'High variance 50': the range of errors is [2,23]
%            Our error histogram       	[ 3 , 14, 14, 9,  4,  2,  1, 1, 0, 2 ]
%            Hwang's error histogram 	[ 0,     3,   7, 9,  7, 12, 6, 3, 2, 1 ]

%Figure \ref{fig:cvpr_compare} shows an accompanying visual comparison. In addition to group C, we also conducted similar experiments on both group D and group E. On images from group D, the mean error of our method on the L channel alone is 7.27, and the mean L2 error on all three channels is 10.97. On group E, our errors are 5.73 and 11.09, respectively. Note that previous work has never reported their performance on groups D and E.
%\begin{figure*}[!t]
%\begin{center}
%   \includegraphics[width=1.0\linewidth]{images/CVPR_2}
%\end{center}
%   \caption{Visual comparison with ~{\protect\cite{Bychkovsky:2011:LPG}}. (a) Input image; (b) groundtruth enhanced image manually produced by expert C; (c) enhanced image by our approach; (d) enhanced image by ~{\protect\cite{Bychkovsky:2011:LPG}}.}
%\label{fig:cvpr_compare}
%\end{figure*}

\begin{figure}[t]
\centerline{\includegraphics[width=1.0\linewidth]{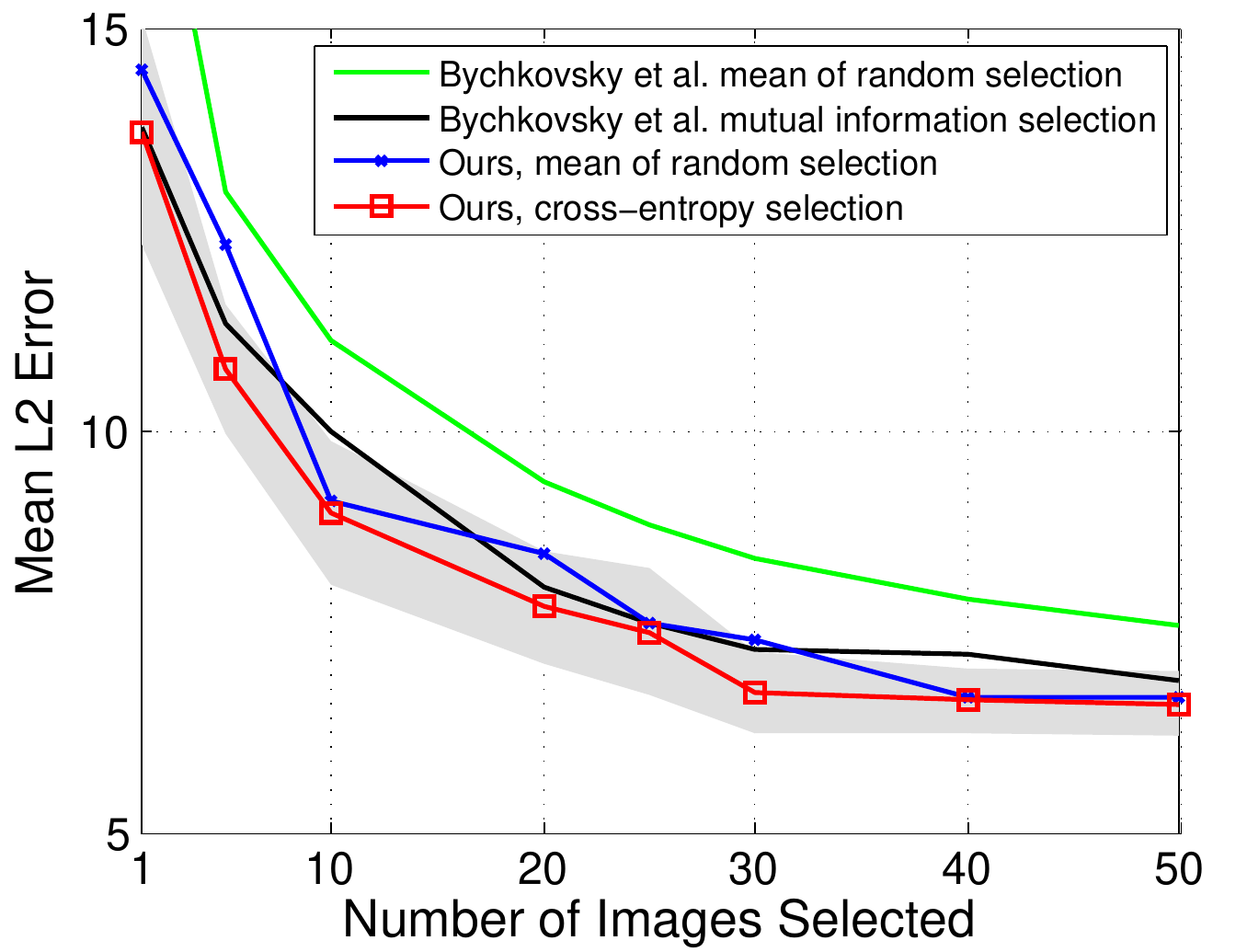}}
   \caption{Comparison of training image selection schemes. When compared with sensor placement based on mutual information, our cross-entropy based method achieves better performance especially when the number of selected images is small. The band shaded in light blue shows the standard deviations of the $L^2$ errors of our scheme.}
\label{fig:cross_entropy_selection}
\end{figure}

To validate the effectiveness of our cross-entropy based training set selection method (Algorithm 1), we have monitored the testing errors by varying the number of training images selected by our method, and compared them with both naive random selection and the sensor placement method used in \cite{Bychkovsky:2011:LPG} (Figure \ref{fig:cross_entropy_selection}). Interestingly, when the random selection scheme is used, our neural network based solution achieves significantly better accuracy than the Gaussian Process based method. This is primarily due to the strong nonlinear regression power exhibited by deep neural networks and the rich contextual feature representation built from semantic analysis. When compared with sensor placement, our cross-entropy based method also achieves better performance especially when the number of selected images is small, which further indicates our method is superior for learning enhancement styles from a small number of training images.

\subsection{Instagram Dataset}
\begin{figure}[!t]
  \centerline{\includegraphics[width=1.0\linewidth]{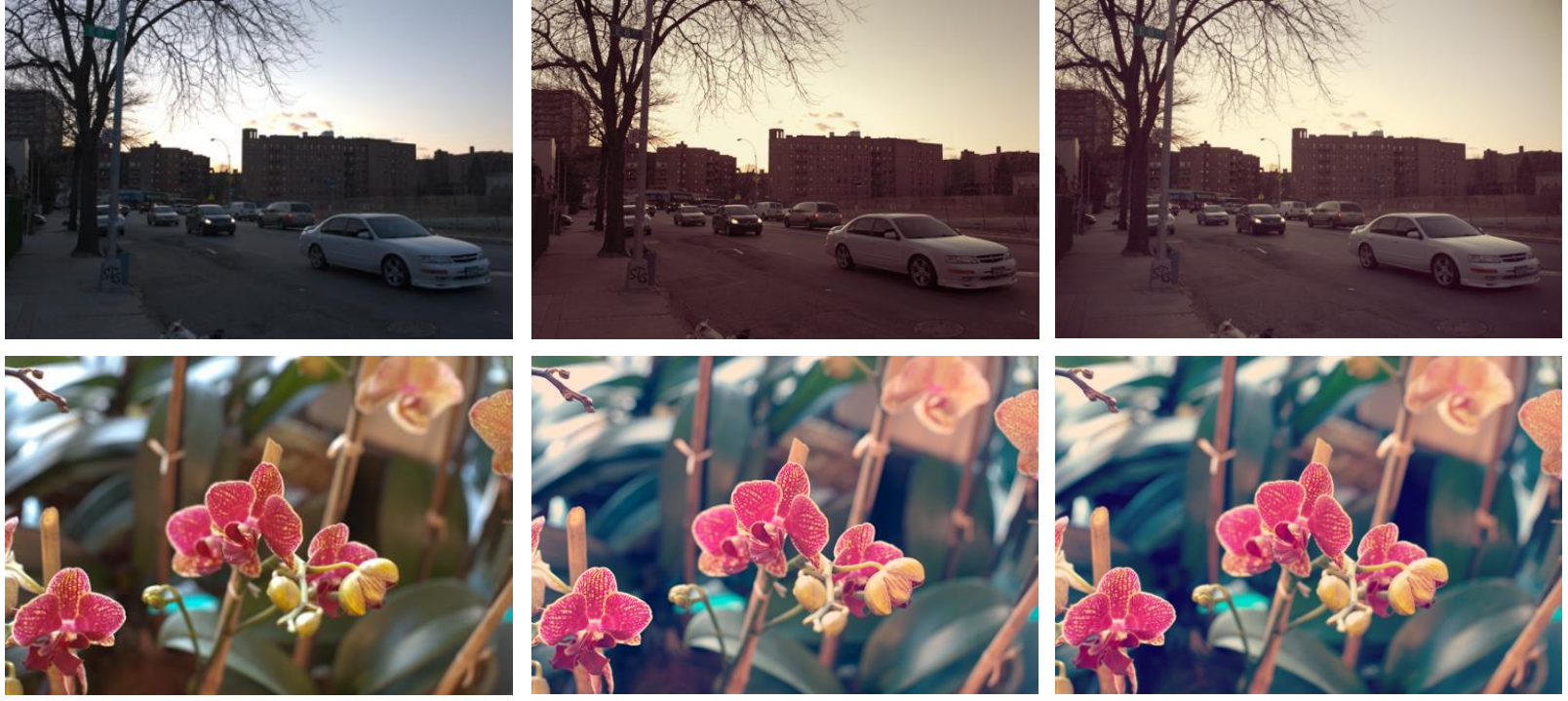}}
  \centerline{\hfill Input Image \hfill\hfill Our Results \hfill\hfill Instagram \hfill}
  \caption{Comparison with Instagram. \textbf{Left}: Input images (from MIT-Adobe FiveK); \textbf{Middle}: our results; \textbf{Right}: results by Instagram. The top row shows the ``EarlyBird" effect, and the bottom row shows the ``Nashville" effect. This comparison indicates enhancement results by our trained color mappings are close to the ground truth generated by Instagram.}\label{fig:Instagram}
\end{figure}

Instagram has become one of the most popular Apps on mobile phones. In Instagram, hundreds of filters can be applied to achieve different artistic color and tone effects. For example, the frequently used ``Lo-Fi" filter boosts contrast and brings out warm tones; the ``Rise" filter adds a golden glow while ``Hudson" casts a cool light. For each specific effect, we randomly chose 50 images from MIT-Adobe FiveK, and let Instagram enhance each of them. Among the resulting 50 pairs of images, half of them were used for training, and the other half were for testing. We have verified whether images adjusted by the trained color mapping functions are similar to the ground truth produced by Instagram, which has the flavor of a reverse engineering task. Our experiments indicate that Instagram effects are relatively easy to learn using our method. Figure \ref{fig:Instagram} shows the learning results for two popular effects.

\subsection{User Studies}\label{sec:user_study}
\begin{figure}[t]
  \centerline{\includegraphics[width=3.2in,height=2.4in]{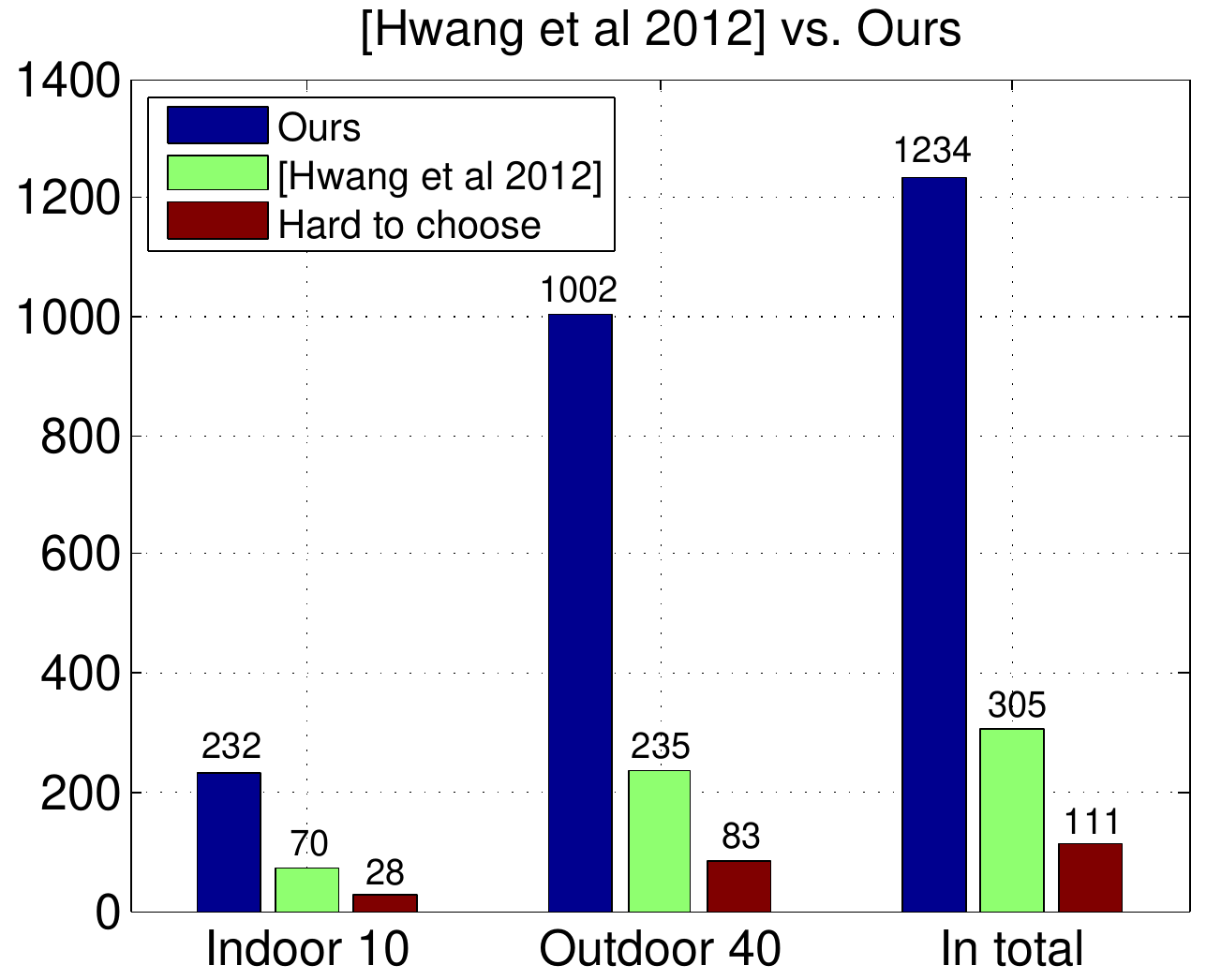}}
  \caption{A comparison of user voting results between our approach and ~{\protect\cite{Hwang:2012:CAL}}.}\label{fig:user_study}
\end{figure}

To perform a visual comparison between our results and those produced by \cite{Hwang:2012:CAL} in an objective way, we collected all the images from the two datasets, ``Random 250" and ``High variance 50", and randomly chose 50, including 10 indoor images and 40 outdoor images, to be used in our user study. For each of these 50 testing images, we also collected the groundtruth images and the enhanced images produced with our method and \cite{Hwang:2012:CAL}. Then we invited 33 participants, including 12 females and 21 males, with ages ranging from 21 to 28. These participants had little experience of using any professional photo adjustment tools but did have experience with photo enhancement Apps such as ``Instagram". The experiment was carried out by asking each participant to open a static website using a prepared computer and a 24-inch monitor with a 1920x1080 resolution. For each test image, we first show the input and the groundtruth image pair to let the participants know how the input image was enhanced by the photographer (retoucher C). Then we show two enhanced images automatically generated with our method and Hwang {\em et al.} in a random left/right layout without disclosing which one was enhanced by our method. The participant was asked to compare them with the ground truth and vote on one of the following three choices: (a) ``The left image was enhanced better", (b) ``The right image was enhanced better", and (c) ``Hard to choose". In this way, we collected 33x50=1650 votes distributed among the three choices. Figure \ref{fig:user_study} shows a comparison of the voting results, from which we can see that enhanced images produced by our method received most of the votes in both indoor and outdoor categories. This comparison indicates that, from a visual perspective, our method can produce much better enhanced images than \cite{Hwang:2012:CAL}.

%The description of our second user study can be found in the supplemental materials.
Our second user study tries to verify whether our method has the capability to enhance a target effect in a statistically significant manner. To conduct this study, we chose 30 test images from one of the local effect datasets described in Section \ref{sec:msr_effect} as our test data. We asked 20 participants from the first study to join our second study. The interface was designed as follows. On top of the screen, we show as the ground truth the enhanced image produced by the photographer we hired, below which we show a pair of images with the left being the original image and the right being the enhanced image produced by our method. Then we asked the participant to assign a score to both the input and enhanced images by considering two criteria at the same time: (1) how closely this image conforms to the impression given by the ground truth, (2) the visual quality of the image. In other words, if the enhanced image looks visually pleasing and closer to the ground truth, it should receive a higher score. For the convenience of the participants, we simply discretized the range of scores into 10 levels. If an image looks extremely close to the ground truth, it should be scored 10. At the end, we collected two sets of scores for the original and enhanced images, respectively. We then conducted the paired T-test on the two sets of scores and found that the two-tail p-value is $p \approx 10^{-10}$, and $t=1.96$, indicating that our approach has significantly enhanced the desired effect from a statistical point of view.

\section{Conclusions and Discussions}
In this paper, we have demonstrated the effectiveness of deep learning in automatic photo adjustment. % We present an automatic technique for complex photo enhancement using deep learning.
We cast this problem as learning a highly nonlinear mapping function
by taking the bundled features as the input layer of a deep neural
network. The bundled features include a pixelwise descriptor, a global
descriptor, as well as a novel contextual descriptor which is built on
top of scene parsing and object detection. We have conducted extensive
experiments on a number of effects including both conventional and
artistic ones. Our experiments show that the proposed approach is able
to effectively learn computational models for automatic
spatially-varying photo adjustment.

\begin{figure}[!t]
\centerline{\includegraphics[width=1.0\linewidth]{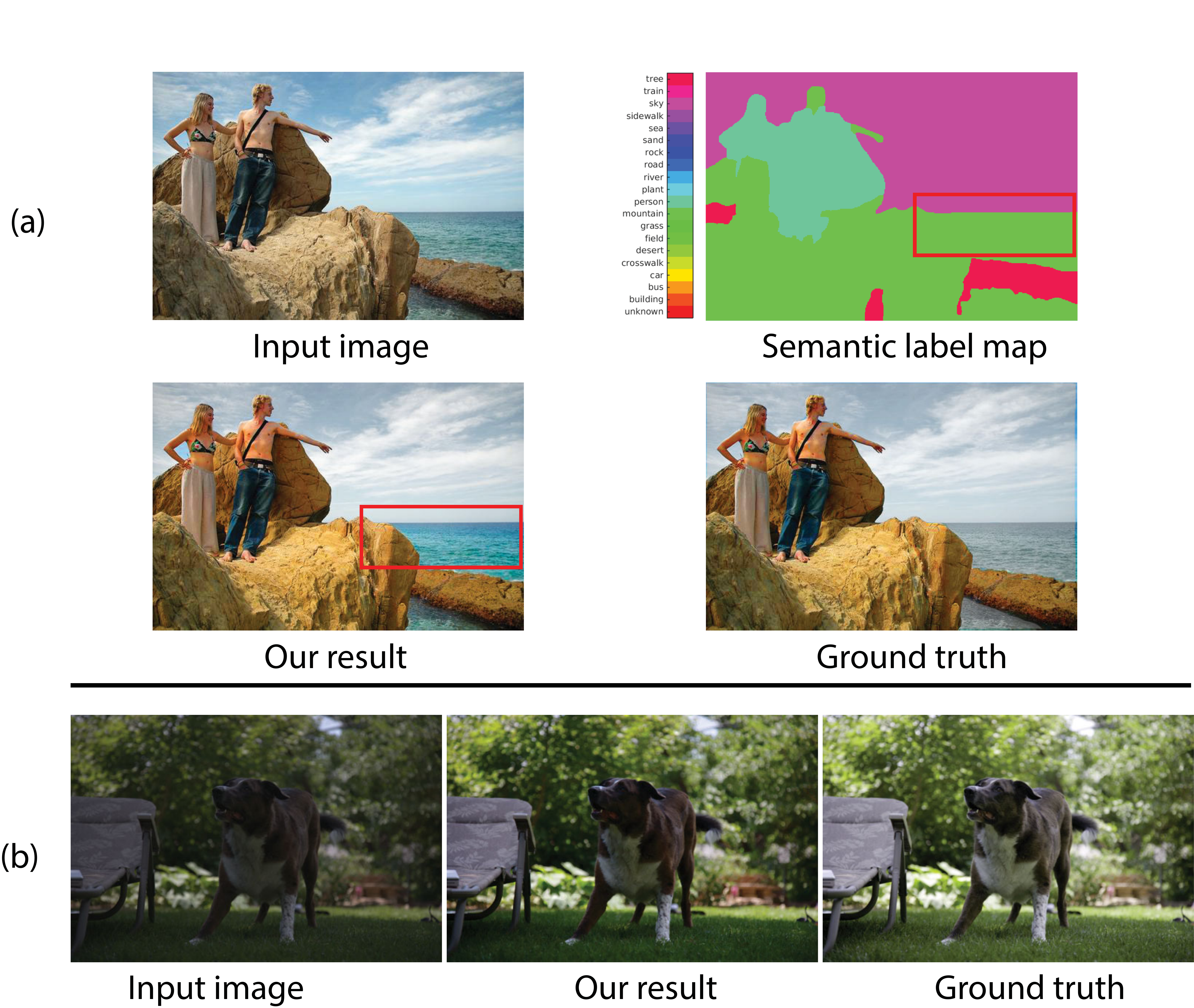}}
  %\centerline{\hfill Input image \hfill\hfill Semantic label map \hfill\hfill Our result \hfill\hfill Ground truth \hfill}
  \caption{Two failure cases. \textbf{Top row}: a failure case on Foreground Pop-Out effect. In the semantic label map, an area with incorrect semantic labeling is highlighted. Correspondingly, this area receives incorrect adjustments in our result. \textbf{Bottom row}: another failure case in "High Variance 50" test set of MIT-Adobe FiveK dataset.}
  \label{fig:failure_case}
\end{figure}

{\flushleft \textbf{Limitations}}. Our approach relies on both scene parsing and object detection to build contextual features. However, in general, these are still challenging problems in computer vision and pattern recognition. \textcolor{black}{Mislabeling in the semantic map can propagate into contextual features and adversely affect photo adjustment. Fig \ref{fig:failure_case}(a) shows one such example for the Foreground Pop-Out effect. The `sea' on the right side is mistakenly labeled as `mountain' and its saturation and contrast are incorrectly increased. As both scene parsing and object detection are rapidly developing areas, more accurate techniques are emerging and could be adopted by our system to produce more reliable semantic label maps.}

\textcolor{black}{Another failure case is shown in Fig \ref{fig:failure_case}(b), where the adjustments in group C of the MIT-Adobe FiveK dataset are learnt. Our method produces insufficient brightness adjustment, which leads to dimmer result than the ground truth. In fact, the $L^2$ distance between the input image and the ground truth is $38.63$, which is significantly higher than the mean distance $17.40$ of the dataset. As our DNN is trained using all available training samples, individual adjustments significantly deviating from the average adjustment for a semantic object type are likely to be treated as outliers and cannot be correctly learnt.}

\textcolor{black}{Our system employs a deep fully connected neural network to regress spatially varying color transforms. There exist many design choices in the DNN architecture, including the number of hidden layers, the number of neurons in each layer, and the type of neural activation functions. They together give rise to a time-consuming trial-and-error process in search of a suitable DNN architecture for the given task. In addition, DNN behaves as a black box and it is not completely clear how the network combines features at different scales and predicts the final color transforms. In fact, interpreting the internal representations of deep neural networks is still an ongoing research topic~\cite{zeiler2013visualizing,szegedy2013intriguing}.}

%Nevertheless, photographers usually pay attention to only a relatively small number of object categories, we can fine-tune the performance of state-of-the-art parsing and detection algorithms on those categories to achieve higher accuracy.

\begin{acks}
We are grateful to Vladimir Bychkovsky and Sung Ju Hwang for fruitful discussions and suggestions.
This work was partially supported by Hong Kong Research Grants Council under General Research Funds
(HKU17209714).
\end{acks}

\bibliographystyle{acmtog}
\bibliography{dnn-img-enhance}

%\received{December 11, 2014}{December 2014}

\end{document}